\documentclass[]{aero}
\usepackage{amssymb}
\usepackage{amsmath}
\usepackage{color,soul}
\usepackage{algorithm}%
\usepackage{algorithmicx}%
\usepackage{algpseudocode}%
\usepackage{subcaption}
\usepackage{booktabs}
\usepackage{graphicx}
\usepackage{xcolor}
\usepackage{tabularx}
\usepackage{booktabs}
\usepackage{array}

\newcommand*{\eg}{\emph{e.g.}}
\newcommand*{\ie}{\emph{i.e.}}
\newcommand*{\etal}{\emph{et al.}}

\ADsetup{
title    = {AI-Enabled Crater-Based Navigation for Lunar Mapping},
author   = {Sofia McLeod$^{1}$\textsuperscript{*}\cor{}, Chee-Kheng Chng$^{1}$\textsuperscript{*}, Matthew Rodda$^{1}$, and Tat-Jun Chin$^{1}$\textsuperscript{$\dagger$}\\ \textsuperscript{*}\textit{Equal contribution.}\\
\textsuperscript{$\dagger$}\textit{SmartSat CRC Professorial Chair of Sentient Satellites.}},
address  = {
  1. The University of Adelaide, Australian Institute for Machine Learning, Corner Frome Road, and North Terrace, Adelaide SA 5000 
},
email    = {sofia.mcleod@adelaide.edu.au},
abstract = {
  Crater-Based Navigation (CBN) uses the ubiquitous impact craters of the Moon observed on images as natural landmarks to determine the six degrees of freedom pose of a spacecraft. To date, CBN has primarily been studied in the context of powered descent and landing. These missions are typically short in duration, with high-frequency imagery captured from a nadir viewpoint over well-lit terrain. In contrast, lunar mapping missions involve sparse, oblique imagery acquired under varying illumination conditions over potentially year-long campaigns, posing significantly greater challenges for pose estimation. We bridge this gap with STELLA - the first end-to-end CBN pipeline for long-duration lunar mapping. STELLA combines a Mask R-CNN-based crater detector, a descriptor-less crater identification module, a robust perspective-n-crater pose solver, and a batch orbit determination back-end. To rigorously test STELLA, we introduce CRESENT-365 - the first public dataset that emulates a year-long lunar mapping mission. Each of its 15,283 images is rendered from high-resolution digital elevation models with SPICE-derived Sun angles and Moon motion, delivering realistic global coverage, illumination cycles, and viewing geometries. Experiments on CRESENT+ and CRESENT-365 show that STELLA maintains metre-level position accuracy and sub-degree attitude accuracy on average across wide ranges of viewing angles, illumination conditions, and lunar latitudes. These results constitute the first comprehensive assessment of CBN in a true lunar mapping setting and inform operational conditions that should be considered for future missions.},
keywords = {Crater-based navigation \sep  Lunar mapping \sep AI-enabled spacecraft localisation \sep Terrain absolute navigation\\
}

}

\begin{document}

\maketitle  




\section{Introduction}
Lunar exploration has gained renewed momentum, with missions targeting a range of objectives including scientific investigation~\cite{blueghost2025, chandrayaan3}, resource mapping~\cite{chin-2024}, and sample return~\cite{nasa_chang_e_6}. These missions span different operational modes, such as flybys, orbiting and landing. Across these diverse mission types, achieving high-precision localisation is critical for mission success.

Vision-based localisation systems are becoming increasingly incorporated in modern space mission design, due in part to the maturation and advancements in camera payloads. Recent advances in computer vision and artificial intelligence have significantly enhanced the capability of vision-based systems, making them increasingly viable for autonomous navigation in complex and variable environments such as the lunar surface.  Moreover, vision-based navigation enables robust localisation when traditional radiometric tracking alone is insufficient.

Terrain absolute navigation (TAN) is a vision-based technique wherein a spacecraft estimates its six degrees of freedom (6DoF) pose, consisting of position and orientation, by matching observed terrain features against a pre-loaded map of the targeted planetary bodies~\cite{Maass2020-to}. TAN has been successfully demonstrated in operational space missions, notably during the OSIRIS-REx sample-return mission~\cite{lorenz2017lessons}, where it enabled precise navigation during proximity operations around the asteroid Bennu. In essence, TAN uses images captured by its onboard camera to identify known surface landmarks, forming 2D–3D correspondences that are then used to estimate its pose relative to the astronomical body~\cite{McLeod2024-ui,johnson2008overview}.

One of the most characteristic terrain features of the lunar surface is its craters. Over the years, lunar crater catalogues have been collated~\cite{Wang2015-gq, Head2010-yq}, with the most recent catalogues detailing over a million craters~\cite{Robbins2019-aw, Wang2021-ab}, providing details on their 3D location and composition. Lunar craters can be parameterised by 3D elliptical discs, which, under perspective projection, are imaged as 2D ellipses. The distinct projective appearance of craters on the image plane and the magnitude of catalogued craters have given rise to crater-based navigation (CBN).

CBN can be applied to various lunar mission scenarios, with prior research focusing largely on lunar landing scenarios~\cite{Maass2020-to, Yu2014-dw, Silvestrini2022-lx, Park2019-aq, Woicke2018-lg, Simard-Bilodeau2012-sz}. While there has been growing interest in lunar mapping missions~\cite{Jeon2024-ob, EhlmannUnknown-cx,chin-2024}, CBN has yet to be fully explored in this context. CBN for lunar landing typically benefits from favourable operational conditions: high image frequency, nadir viewpoints, and well-lit local terrain. Comparatively, lunar mapping missions impose broader and more challenging constraints.  Orbiters may capture only one image every few minutes, or even hours, while mapping the entire Moon over a multi-year campaign. These images may be acquired from oblique viewpoints and under varying illumination conditions, making the pose estimation task significantly more challenging than in landing scenarios. A comparison of these two mission profiles is summarised in Tab.~\ref{tab:landing_vs_surveillance}, highlighting the fundamental differences that motivate the need for a dedicated CBN pipeline tailored to lunar mapping. In this work, we address this gap by designing a CBN pipeline for lunar mapping that considers, and is evaluated under all conditions outlined in Tab.~\ref{tab:landing_vs_surveillance}.

\begin{table}[h!]
\centering
\caption{Comparison between lunar landing and lunar mapping scenarios.}
\begin{tabular}{|l|p{5cm}|p{5cm}|}
\hline
\textbf{Aspect} & \textbf{Lunar landing} & \textbf{Lunar mapping} \\
\hline
Lighting condition & Typically well-lit to ensure safe descent and visibility of terrain features. & A diverse range of lighting conditions, including low-angle and shadowed regions. \\
\hline
Crater catalogue & A small, high-resolution catalogue focused on the target landing site. & A complete catalogue covering the entire lunar surface. \\
\hline
Camera-pointing angle & Often assumed to be nadir. &  May include off-nadir angles. \\
\hline
Mission campaign & Typically in the magnitude of hours or minutes, encompassing the entire descent duration. & In the magnitude of months or years to achieve global lunar mapping coverage. \\
\hline
Imaging frequency & Potentially high imaging frequency to enable frequent pose estimates during the landing sequence. & Typically low imaging frequency due to downlinking restrictions and resource constraints.\\
\hline
\end{tabular}
\label{tab:landing_vs_surveillance}
\end{table}

\subsection{CBN pipeline}~\label{subsec:CBN_pipeline}
\begin{figure}[H]
     \centering
     \includegraphics[width=0.9\linewidth]{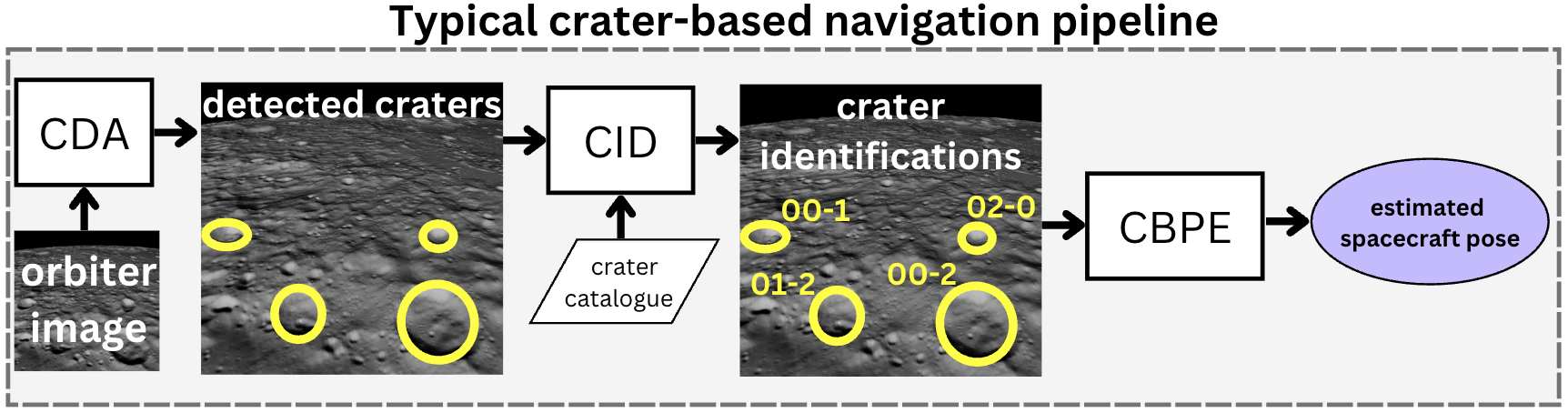}
     \caption{A typical CBN pipeline.}\label{fig:cbn_pipeline}
\end{figure}

A typical CBN pipeline entails a crater detection algorithm (CDA)~\cite{Woicke2018-lg, Downes2020-ra, Gnam2025-ol}, a crater identification (CID) algorithm~\cite{Christian2021-ts, Hanak2010-fp, Park2019-aq} and a final crater-based pose estimation (CBPE) algorithm~\cite{Lu2016-bl, Wokes2010-nv, Chen2021-ai}. From an image of the observed lunar surface provided by an onboard optical sensor, the CDA detects the craters in the image, often fitting an ellipse to these detections, parameterising the crater rims. CID then matches these detected craters to known craters in the crater catalogue, and CBPE uses these crater identifications to estimate the pose of the spacecraft. A depiction of this pipeline can be seen in Fig.~\ref{fig:cbn_pipeline}. The following sections review the literature on the mentioned components.


\subsubsection{Crater detection algorithm}
CDAs are a mature area of research, with applications ranging from automated crater detection tools used to expand existing crater catalogues~\cite{Lee2019-sl, Benedix2020-ru}, to detecting craters for CBN pipelines~\cite{Maass2020-to, McLaughlin2022-wq, Silvestrini2022-lx}. While these applications all require their employed CDA to detect craters, these CDAs may not necessarily be cross-compatible. Lee and Benedix~\etal~both proposed CDAs for Martian crater catalogue expansion. However, their methods are not directly transferable to a CBN context as they operate on non-optical imagery (\eg, digital elevation models (DEM) or thermal-infrared images)~\cite{Lee2019-sl, Benedix2020-ru}.

Significant progress has been made in CDAs developed using machine learning approaches, where they have exhibited superior performance over classical algorithms~\cite{Christian2021-ts, Woicke2018-lg}. Woicke~\etal~demonstrated the superior performance of a supervised learning-based CDA over five handcrafted CDAs~\cite{Woicke2018-lg}. Meanwhile, Downes~\etal~proposed LunarNet, a convolutional neural network (CNN)-based CDA, displaying robustness to noise and illumination conditions in lunar images~\cite{Downes2020-ra}. LunarNet reinforces the superior performance of deep learning-based approaches over traditional image processing techniques. Additionally, they demonstrated that knowledge learned from DEMs can be transferred to optical images. 

Despite these advancements, data availability remains a major bottleneck for training deep learning-based CDAs. There are relatively few real lunar image datasets suitable for supervised learning. Notably, datasets collected by the Lunar Reconnaissance Orbiter (LRO) Narrow-Angle Camera (NAC) and Wide-Angle Camera (WAC)~\cite{Robinson2010-dd} are primarily composed of nadir-viewing images captured under well-lit conditions, limiting their usefulness for training models intended for more varied lunar mapping scenarios. This lack of diversity in viewpoint and lighting reduces the generalisation capability of models trained solely on such data.

To overcome this, researchers have turned to synthetic data. Synthetic datasets generated using modelling tools like PANGU~\cite{Silvestrini2022-lx, Woicke2018-lg} offer a flexible way to define a wide range of environmental and operational parameters, including spacecraft altitude, solar incidence angle, imaging frequency, and crater size and distribution. Silvestrini~\etal~\cite{Silvestrini2022-lx} and Woicke~\etal~\cite{Woicke2018-lg} used synthetic images for both training and evaluation, enabling controlled experiments across diverse scenarios. However, because these simulations rely on synthetically generated DEMs that do not reflect real lunar topography, the resulting imagery lacks the geological realism needed for high-fidelity evaluation.

To improve realism, more recent work has explored rendering lunar images using real DEMs collected during previous lunar missions~\cite{Ohtake2008-cq, lro_dataset-mit}. For instance, Simard Bilodeau~\etal~\cite{Simard-Bilodeau2012-sz} simulated landing trajectories over DEM-derived surfaces (from the Kaguya mission~\cite{Ohtake2008-cq}), producing synthetic images that better capture the geometric and visual characteristics of the actual lunar environment. While their dataset more closely resembled real lunar surface conditions compared to Silvestrini~\etal~\cite{Silvestrini2022-lx} and Woicke~\etal~\cite{Woicke2018-lg}, it was still limited by the low resolution of the available DEMs, necessitating the artificial addition of some craters. Furthermore, their evaluation demonstrated that crater detection performance degraded significantly under non-ideal lighting and oblique viewing conditions.

Building on these efforts, McLeod~\etal proposed CRESENT~\cite{lro_dataset-mit}, a large-scale synthetic dataset comprising over 5,000 images generated using high-resolution DEMs from the more recent LOLA mission. CRESENT features a wide range of viewing angles and lighting conditions, offering a richer training environment for crater detection models. More recently, Rodda~\etal~\cite{Rodda2024-oz} investigated the domain gap between synthetic imagery (CRESENT) and real lunar imagery captured during the Chang'e~5 mission~\cite{cnsachange5_2020}. They manually labelled craters in the Chang'e~5 images to construct a real-world dataset, CDA-CE5, which consists of off-nadir images captured sequentially in a landing sequence. Their study revealed that a Mask R-CNN-based CDA trained solely on CRESENT performed poorly when evaluated on CDA-CE5, but fine-tuning the model on CDA-CE5 data substantially improved performance. While their analysis provided important insights into domain transferability, it did not address the impact of crater detection performance on the downstream stages of a complete CBN pipeline. Motivated by this gap, we adopt and retrain the model proposed by~\cite{Rodda2024-oz} on a larger dataset, CRESENT+~\cite{McLeod2025} (details in Sec.~\ref{sec:cresent}), and systematically evaluate its influence across the entire CBN pipeline, including the latest CID and CBPE modules.

\subsubsection{Crater identification}
CID has been studied extensively over the last few decades. The mainstream approach of CID is based on descriptor matching, \ie, the descriptors extracted from detected craters are used to query the closest match from a descriptor database constructed with the catalogued craters. Much of the prior work has focused on exploring robust crater descriptors. For instance, Cheng and Ansar~\cite{cheng2005landmark} proposed to extract conic invariants from crater pairs, while Hanak~\etal~\cite{hanak2009lost, hanak2010crater} proposed descriptors based on the triangle formed by the centre points of crater triads. 

More recently, Christian~\etal~\cite{Christian2021-ts} highlighted that most existing descriptors fail to represent the crater groups with the maximum information. Leveraging invariant theory, they introduced a set of descriptors that maximise the number of algebraically independent and permutation-invariant invariants for crater combinations.

Chng~\etal~\cite{chng2024crater} further argued that most of the existing descriptors rely on restrictive assumptions about the underlying surface geometry. Common assumptions include a nadir-pointing camera~\cite{hanak2009lost, hanak2010crater}, coplanar crater arrangements~\cite{cheng2005landmark, Park2019-aq}, and circular crater shapes~\cite{Maass2020-to}. While these assumptions are often valid for landing missions—where the camera typically looks straight down at a small, planar region—they are less applicable to more general scenarios. As shown in~\cite{chng2024crater}, when these assumptions are violated, CID performance deteriorates significantly. To address this limitation, Chng~\etal~\cite{chng2024crater} proposed PECAN, the first descriptor-less CID framework. Unlike prior methods, PECAN makes no assumptions about the imaging geometry or crater configuration, and it achieves state-of-the-art performance across a diverse set of testing conditions.

Previous studies on CID have typically relied on small mission-specific catalogues containing only a few thousand craters~\cite{Christian2021-ts, hanak2009lost, chng2024crater}. This approach was motivated by the need to improve descriptor uniqueness and reduce memory consumption. However, we found that limiting the mission catalogue size can lead to missed correspondences and degraded performance in subsequent pose estimation stages. To overcome this limitation, we instead employ a much larger mission catalogue based on the full Robbins dataset~\cite{Robbins2019-aw}, comprising over one million craters. To the best of our knowledge, this is the first time a CID method has been evaluated at this scale, revealing new insights into the benefits of large-scale mission catalogues for robust lunar mapping navigation.

\subsubsection{Crater-based pose estimation}

\begin{figure}[H]
     \centering
     \includegraphics[width=0.7\linewidth]{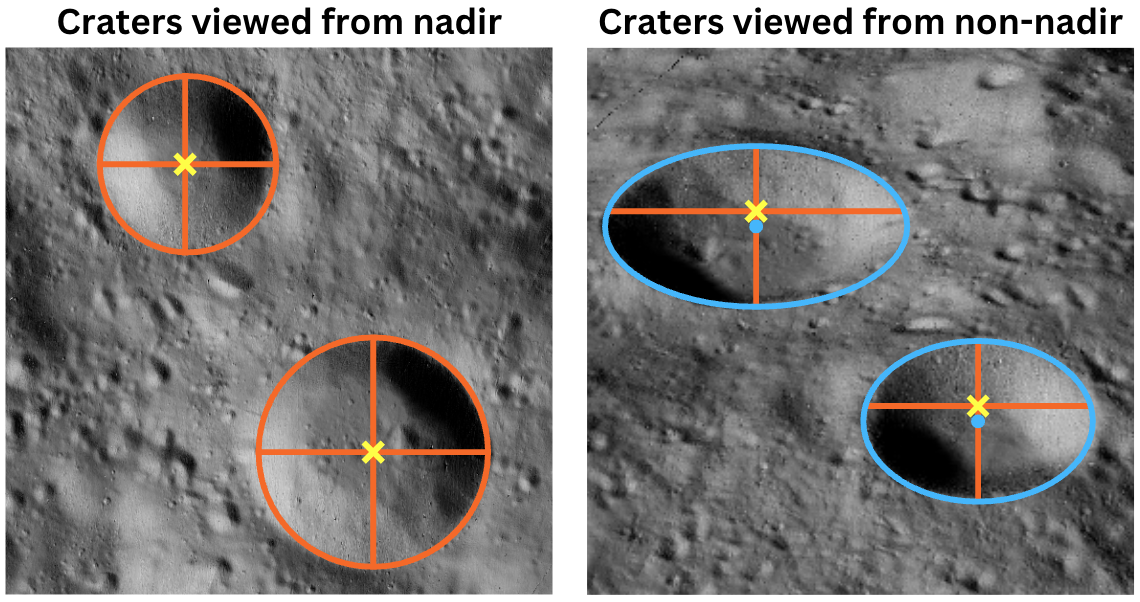}
     \caption{The centre of a crater disc (yellow cross) does not generally align with the centre of the imaged crater rim ellipse (blue ellipse) under non-nadir observation.}\label{fig:nadir_non_nadir_craters}
\end{figure}

From the correlation between the 2D detected craters and the 3D catalogued craters made by CID, the pose of the spacecraft can be estimated through a CBPE algorithm. CBPE has had considerably less development in the CBN literature, with many pipelines like the ones presented by Chen and Jiang, Xu~\etal, and Maass~\etal~estimating spacecraft pose through variations of the perspective-n-point (PnP) algorithm~\cite{Chen2021-ai, Xu2022-jf, Maass2020-to}.  CBN pipelines that employ PnP typically reduce the information-rich, elliptical representation of detected and catalogued craters to single points representing the crater centres~\cite{Christian2021-ts}. In many cases, the 2D projection centre of the detected crater is represented by the ellipse centre, which only holds under nadir observation (see Fig.~\ref{fig:nadir_non_nadir_craters}). 

A recent study demonstrated that a CBPE optimisation method that utilised the elliptical representation of craters rather than points, yielded significantly higher pose estimation accuracy and was robust to non-nadir viewing angles, simulated crater detection noise and irregular terrain~\cite{McLeod2025}. This ellipse-based perspective-n-crater (PnC) CBPE method proposed by McLeod~\etal~has been evaluated on the CRESENT dataset~\cite{McLeod2024-ui} and the CRESENT+ dataset~\cite{McLeod2025}, both of which emulate some of the expected conditions of a lunar mapping mission~\ie, low lunar orbit altitude of 100km and a variety of camera viewing angles. While PnC has demonstrated superior performance over other optimisation-based CBPE methods, it has not been evaluated in a complete CBN pipeline.

\subsection{Contributions}

In this work, we propose a new CBN pipeline, \textbf{STELLA} - \textbf{S}pacecraft cra\textbf{T}er-bas\textbf{E}d \textbf{L}ocalisation for \textbf{L}unar m\textbf{A}pping. To the best of our knowledge, STELLA is the first CBN system designed and evaluated under the expected conditions of a lunar mapping mission. 

The proposed STELLA pipeline comprises two main components: STELLA-core and STELLA-OD. STELLA-core is invoked when a new image is acquired, performing CDA, CID, and CBPE in a sequential fashion, returning a pose estimate for each image when possible. Meanwhile, STELLA-OD is run when a batch of position estimates is collected from STELLA-core. It performs orbit determination (OD) to get an orbit that best aligns with the position estimates in a least-squares sense. The orbit can then be propagated to the timestamps where STELLA-core fails to yield an output.

To provide a comprehensive evaluation of the STELLA pipeline under the expected conditions of a lunar mapping mission (see Tab.~\ref{tab:landing_vs_surveillance}), we introduce a new benchmarking dataset, CRESENT-365. The CRESENT-365 dataset was constructed to closely emulate the imaging frequency, viewing geometry, and surface illumination variations expected in an actual lunar mapping mission. It contains 15,283 images that provide a near-global coverage of the lunar surface. To the best of our knowledge, CRESENT-365 is the most extensive and realistic lunar mapping imaging dataset to exist, and will be publicly released as a further contribution of this work. Benchmarking the STELLA pipeline on both the CRESENT+ and CRESENT-365 datasets demonstrated its robust performance, even under challenging illumination conditions and viewing geometries, highlighting its suitability for lunar mapping missions.

\section{Problem context}\label{sec:problem_context}
In this section, we outline the problem context of CBN for lunar mapping missions, defining the relevant reference frames and camera projection model.

\subsection{CBN for lunar surface mapping}\label{sec:mission_specs}

Lunar mapping missions aim to identify and map lunar surface resources from sensor measurements, where the location of the sensor measurements on the lunar surface can be obtained, provided the pose of the spacecraft is known. To build a complete map of lunar resources, we assume that the spacecraft maintains a low lunar orbit (approximately 100km altitude) over a long campaign (in the magnitude of months or years) to achieve near-global surface coverage. While a rough pose estimate of the spacecraft may be provided through external sensors,~\eg, star trackers and radio ranging~\cite{McLeod2024-ui}, this has to be refined to achieve the localisation accuracy required by the mission. Therefore, we propose a CBN system to estimate spacecraft pose for lunar mapping mission applications.

\begin{figure}[H]
     \centering
     \includegraphics[width=0.5\linewidth]{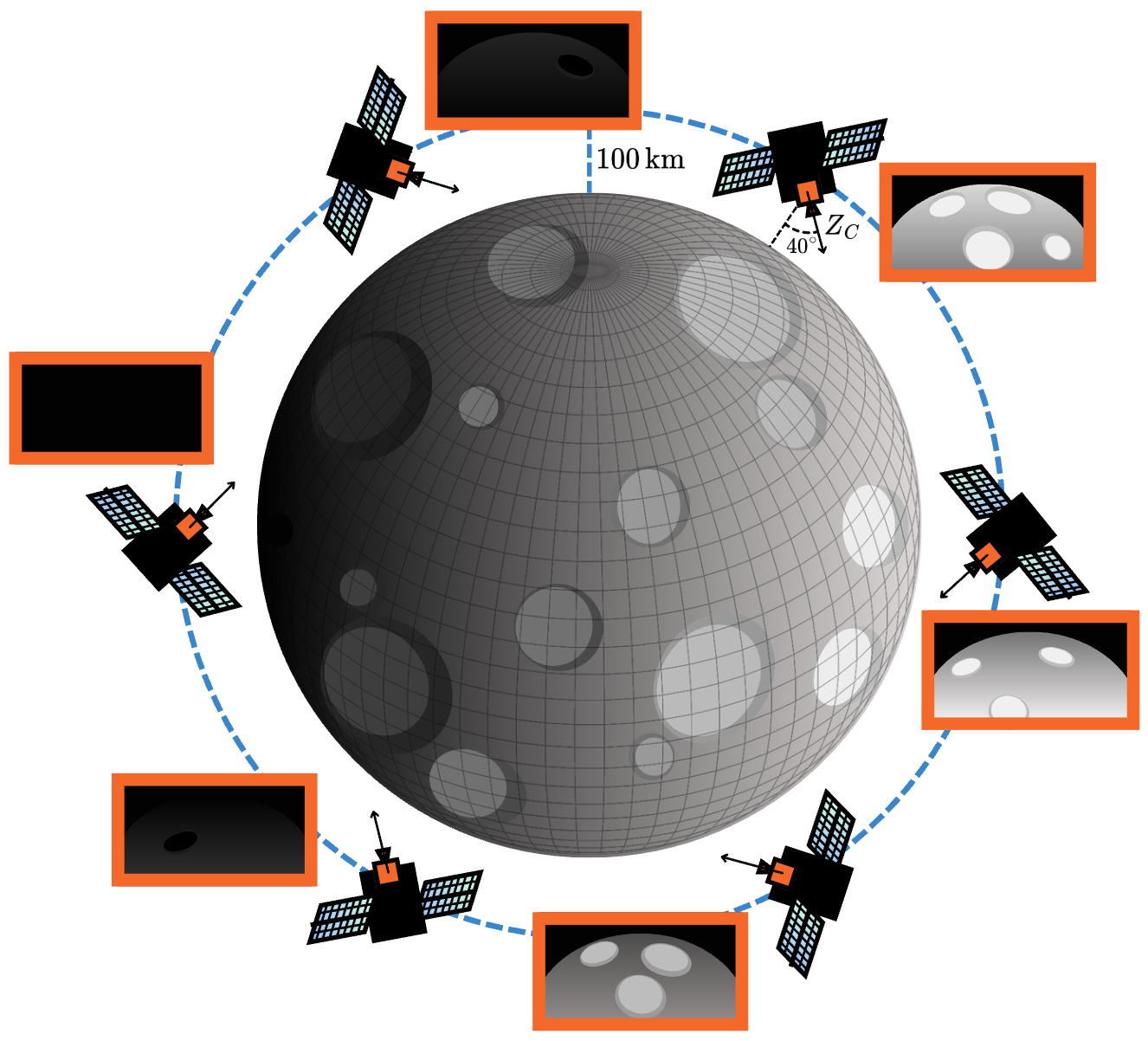}
     \caption{Visualisation of a lunar mapping mission following a low lunar orbit trajectory, with an optical sensor imaging the surface at a rate of six images per orbit at an angle of $40^\circ$ off nadir.}\label{fig:orbital_trajectory}
\end{figure}

In this work, we assume an optical sensor onboard a spacecraft will operate over a long-duration mission with a fixed frequency, imaging the lunar surface under various illumination conditions, without assuming a nadir-pointing camera (see Fig.~\ref{fig:orbital_trajectory}).  In Sec.~\ref{sec:CRESENT-365}, we present a dataset, CRESENT-365, which consists of images taken under these conditions, produced under a simulated lunar mapping mission.

From the images captured by the optical sensor during the mission, a CBN pipeline can be employed to estimate the pose of the spacecraft.  The following section details the reference frames considered in the context of CBN for lunar mapping missions.

\subsection{Reference frames}

To understand the relationship between the pose of the spacecraft and the craters imaged by the optical sensor, we must first know the reference frames considered for CBN.

\begin{figure}[H]
     \centering
     \includegraphics[width=0.5\linewidth]{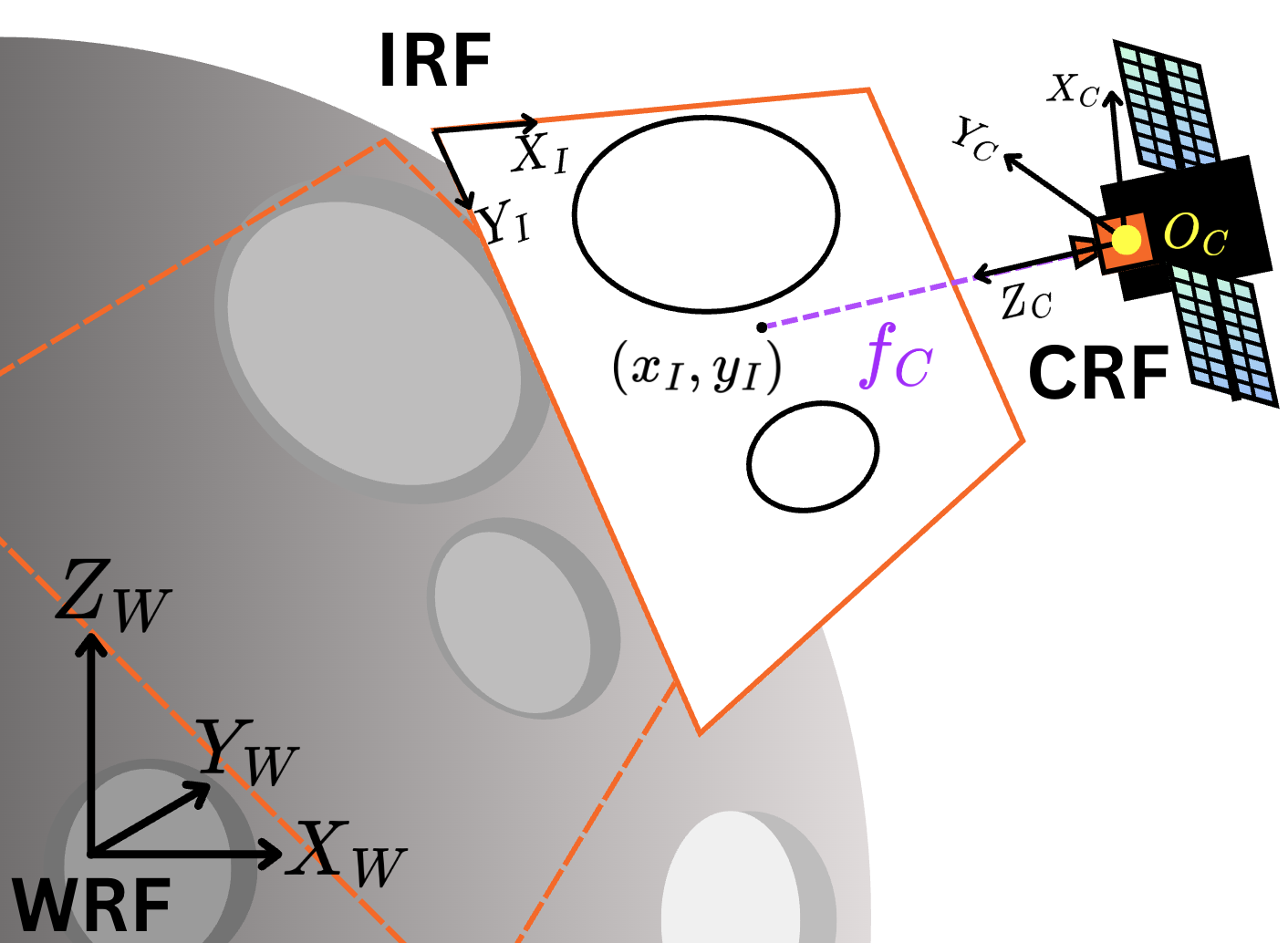}
     \caption{The WRF, IRF and CRF considered in this work.}\label{fig:reference_frames}
\end{figure}

Craters on the lunar surface are specified in relation to the Selenographic (Moon-centred, Moon-fixed) reference frame, which we will refer to as the world reference frame (WRF). Without loss of generality, the spacecraft reference frame (SRF) is aligned with the camera reference frame (CRF), where the view direction of the camera is along the $Z_C$ axis of the CRF. The image reference frame (IRF) is located at a focal length distance, $f_C$, from the origin $O_C$ of the camera, where the origin of the IRF is at point $(x_I, y_I)$. A depiction of these reference frames can be seen in Fig.~\ref{fig:reference_frames}.  We highlight that the goal of CBN is to estimate the pose of the spacecraft in the WRF by finding the transformation from the CRF to the WRF.  This transformation is detailed in the following section.

\subsection{Camera projection model}

The projection matrix of a camera provides a way of transforming points defined in the WRF to corresponding points in the CRF and IRF.  It consists of both camera intrinsic and extrinsic matrices. 

In this work, we assume a calibrated camera with no distortion.  The camera intrinsic matrix can therefore be given by,
\begin{equation}\label{eq:intrinsic_matrix}
    \mathbf{K} = \begin{bmatrix}
        f_C & 0  & x_I \\
        0 & f_C & y_I\\
        0 & 0 & 1
    \end{bmatrix}.
\end{equation}
The camera extrinsic matrix, $\mathbf{J}^W_C$, describes the transformation from CRF to WRF,~\ie, the 6DoF pose of the spacecraft in the WRF.  The pose parameters constitute a position vector, $\mathbf{t}_W \in \mathbb{R}^{3\times1}$ specified in the WRF, and a rotation matrix that exists in the special orthogonal group, $\mathbf{T}^W_C\in SO(3)$, from the CRF to the WRF (which can be constructed from three Euler angles). From this, $\mathbf{J}^W_C$ is defined as,
\begin{equation}\label{eq:extrinsic_matrix}
    \mathbf{J}^W_C = \mathbf{T}^W_C[\mathbf{I}_{3 \times 3} | -\mathbf{t}_W] = [\mathbf{T}^W_C | \mathbf{t}_C],
\end{equation}
\noindent
where $\mathbf{I}$ is a $3\times 3$ identity matrix.

Using both camera intrinsic and extrinsic matrices, the camera projection matrix can be defined as,
\begin{equation}\label{eq:projection_matrix}
    \mathbf{P}^W_C = \mathbf{K} \cdot \mathbf{J}^W_C.
\end{equation}
\noindent
This matrix can project homogeneous point $\mathbf{\bar{q}}_W$ in the WRF, to a point $\mathbf{q}_C$ in the CRF through the following transformation,
\begin{equation}\label{eq:project_point_to_CRF}
    \mathbf{q}_C = \mathbf{P}^W_C\cdot\mathbf{\bar{q}}_W,
\end{equation}
\noindent
and then to point $\mathbf{q}_I$ in the IRF by,
\begin{equation}\label{eq:project_point_to_IRF}
    \mathbf{q}_I = \left( \dfrac{\mathbf{q}_C[1]}{\mathbf{q}_C[3]}, \dfrac{\mathbf{q}_C[2]}{\mathbf{q}_C[3]}  \right),
\end{equation}
\noindent 
where $\mathbf{q}_C[i]$ accesses the $i^{th}$ element of $\mathbf{q}_C \in \mathbb{R}^{3 \times 1}$.

This projection matrix also allows us to perform perspective projection on more complex geometric objects, like conics describing crater rims.  The following section describes the perspective projection of a crater rim conic - a fundamental concept for CBN.

\subsubsection{Projection of a crater}

A crater in the WRF can be parameterised by a conic matrix, $\mathbf{C}$, describing the elliptical disc of the crater rim.  Following~\cite{Christian2021-ts}, the crater disc will have a local East-North-Up (ENU) reference frame, $\mathbf{N}_E$, originating at the 3D centre point of the disc, $\mathbf{c}_W$.  From this, a 2D homogeneous point, $\mathbf{\bar{d}}_E$, on the crater rim in the ENU reference frame can be transformed to a homogeneous 3D point $\mathbf{\bar{d}}_W$ in the WRF by, 
\begin{equation}\label{eq:ENU_point_to_WRF_point}
    \bar{\mathbf{d}}_W = \begin{bmatrix} \mathbf{B} \\ \mathbf{k}^T\end{bmatrix} \cdot \bar{\mathbf{d}}_E ,
\end{equation}
\noindent where,
\begin{equation}\label{eq:ENU_point_to_WRF_point_parameters}
    \mathbf{B}=\begin{bmatrix}\mathbf{N}_E \cdot \mathbf{S} \;\;\;\; \mathbf{c}_W\end{bmatrix} , 
    \;\;\;\;
    \mathbf{S} = \begin{bmatrix}
        \mathbf{I}_{2 \times 2} \\ \mathbf{0}_{1 \times 2} 
    \end{bmatrix} ,
    \;\;\;\; \mathbf{k} = [0, 0, 1]^T .
\end{equation}
\noindent
The homography $\mathbf{H}$ between the local disc plane of the crater and the image plane of the camera can therefore be described as,
\begin{equation}\label{eq:homography}
    \mathbf{H} = \mathbf{P}^W_C \begin{bmatrix}
        \mathbf{B} \\ \mathbf{k}^T
    \end{bmatrix}.
\end{equation}
\noindent
For brevity, we refer to~\eqref{eq:homography} as function $h$, \ie, $\mathbf{H} = h(\mathbf{T}^W_C, \mathbf{t}_W, \mathbf{K})$, henceforth. 
From this, we obtain the projection of the locally defined crater conic $\mathbf{C}$ to a conic $\mathbf{E}$ on the image plane through the following relationship,
\begin{equation}\label{eq:conic_projection_expanded}
    \mathbf{E}^{-1} \propto \mathbf{H} \cdot \mathbf{C}^{-1} \cdot \mathbf{H}^T .
\end{equation}
Through a function $c$, the projected conic matrix can be converted to the 2D ellipse parameters, \ie, $\mathcal{E} = \{x, y, \hat{a}, \hat{b}, \theta\}$, where $(x,y)$ denotes the ellipse centre coordinates in the IRF, $\hat{a}$ and $\hat{b}$ are the lengths of the semi-major and semi-minor axes, and $\theta$ is the ellipse orientation with respect to the IRF. We denote the function compactly as below,
\begin{equation}\label{eq:conic_to_ellipse_convert}
    \mathcal{E} = c(\mathbf{E}),
\end{equation}
\noindent and leave the full expression in Sec. 1 of the supplementary material. For notational convenience, we collect the operations in~\eqref{eq:homography} to~\eqref{eq:conic_to_ellipse_convert} into the composite mapping, 
\begin{align}\label{eq:conic_to_ellipse}
\begin{aligned}
    \mathcal{E} &= v(\mathbf{C}, \mathbf{T}^W_C, \mathbf{t}_W, \mathbf{K}) \\
                &:=c(h(\mathbf{T}^W_C, \mathbf{t}_W, \mathbf{K}) \: \mathbf{C}^{-1} \: h(\mathbf{T}^W_C, \mathbf{t}_W, \mathbf{K})^{T}) .
\end{aligned}
\end{align}
\noindent
In this section, we have established the projective relationship between crater rims defined in the WRF and their imaged correspondences under camera pose $(\mathbf{T}^W_C, \mathbf{t}_W)$. The functions detailed here are integral to CBN, especially in the CID and CBPE components, which will be detailed in Sec.~\ref{sec:cid} and Sec.~\ref{sec:cbpe}.

Before introducing our proposed CBN pipeline in Sec.~\ref{sec:stella}, the following section details the two datasets curated for this work. These datasets encompass the expected conditions of a lunar mapping mission, motivating the design of the CBN pipeline.

\section{Datasets}\label{sec:datasets}
In this work, we utilise two different lunar datasets. Our CDA was trained on the existing CRESENT+~\cite{McLeod2025} dataset, and our STELLA pipeline was evaluated on both a testing set of CRESENT+, and a new lunar mapping dataset, CRESENT-365.  In this section, we review CRESENT+, introduce CRESENT-365, and highlight the crater catalogue used in this work.

\subsection{CRESENT+}\label{sec:cresent}
CRESENT+ is an existing image dataset, emulating some of the expected conditions of a lunar mapping mission~\cite{McLeod2025}.  Using PANGU, a state-of-the-art planetary surface simulation software, images were rendered at an altitude of 100km above one of four high-resolution lunar DEMs~\cite{lro_dataset-mit}, each with latitude and longitude dimensions of $45^\circ$.  The original aim of CRESENT+ was to provide an extensive dataset of well-illuminated lunar surface images at viewing angles ranging from $20^\circ-65^\circ$ off-nadir, in increments of $5^\circ$.  As CRESENT+ provides the ground truth pose of each image, ground truth imaged crater locations can be obtained by projecting craters from a crater catalogue (see Sec.~\ref{sec:crater_catalogue}).

We trained our CDA exclusively on the CRESENT+ dataset to assess its performance when exposed to unseen and more challenging lunar terrains presented in CRESENT-365. Since CRESENT+ consists primarily of images captured under well-illuminated conditions, we applied data augmentation during training to simulate both extremely bright and dark imaging scenarios (see Sec.~\ref{sec:cda}). In parallel, evaluating on the CRESENT+ testing set allows us to specifically examine the behaviour of the proposed pipeline under varying oblique viewing angles, providing insight into its suitability for lunar mapping missions.

\subsection{CRESENT-365}\label{sec:CRESENT-365}
To evaluate the performance of our proposed CBN pipeline under realistic lunar mapping conditions, we introduce a new dataset: \textbf{CRESENT-365}. To simulate a representative trajectory, we define a nearly circular, polar orbit at an altitude of 100km with the parameters defined in Tab.~\ref{tab:orbital_parameters}.

\begin{table}[h]
\centering
\caption{Keplerian orbital parameters for the CRESENT-365 dataset}
\label{tab:orbital_parameters}
\begin{tabular}{ll}
\hline
\textbf{Parameter} & \textbf{Value} \\
\hline
Semi-major axis ($a$) & 1837.7 km \\
Eccentricity ($e$) & $3.8e^{-4}$ \\
Inclination ($i$) & $90^\circ$ \\
Longitude of the Ascending Node ($\Omega$) & $227^\circ$ \\
Argument of Periapsis ($\omega$) & $295^\circ$ \\
True Anomaly ($\nu$) & $91^\circ$ \\
\hline
\end{tabular}
\end{table}

The simulated mission is initiated at 00:00:00 on February 20, 2025, with a low-frequency imaging rate of one image every 1200 seconds, yielding 6 images per orbit. The spacecraft is propagated using a Two-body Keplerian model~\cite[Chap.2]{vallado2001fundamentals}. The onboard camera is oriented at a fixed $40^\circ$ off-nadir angle with respect to the radial vector and is tilted in the along-track direction, enabling a forward-looking perspective.

Over a full year of operation (365 Earth days), this setting results in a large number of spacecraft poses, including camera poses that would capture the non-illuminated parts of the Moon. We filter out those poses by using the solar angle to the surface,
\begin{equation}\label{eqn:solar_angle}
    \phi = \angle(\mathbf{n}, \mathbf{s}),
\end{equation}
where $\angle(,)$ denotes the angular distance computation, $\mathbf{n} \in \mathbb{R}^3$ is the normalised intersection point between the principal angle of the camera and the lunar surface, and $\mathbf{s} \in \mathbb{R}^3$ is the direction of the Sun defined in the coordinates of the Moon. We filtered out poses that yield $\phi$ that is greater than $90^\circ$. After the filtering process, 15,283 poses remained, averaging more than three poses per orbit\footnote{Average: 3.48; median: 3; minimum: 3; maximum: 5}.

\begin{figure}[!hbt]
    \centering
    \includegraphics[width=0.7\linewidth]{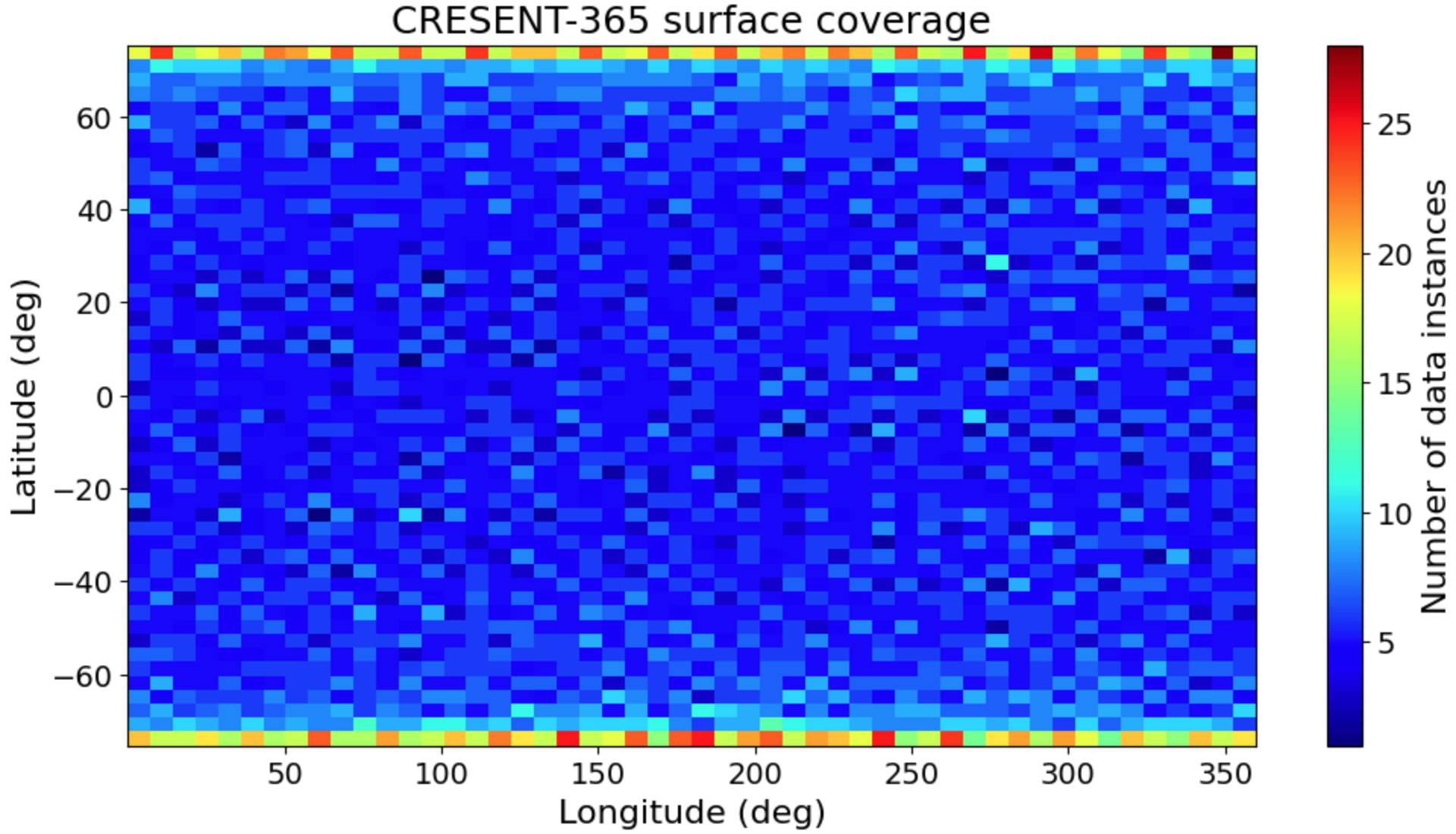}
    \caption{Number of CRESENT-365 data instances visualised across the lunar surface.}
    \label{fig:map_coverage}
\end{figure}

These poses were then used to simulate realistic lunar surface images using PANGU. Given the timestamped trajectory and associated camera poses, PANGU synthesises high-fidelity images using DEMs provided by the LOLA mission, hosted at MIT~\cite{lro_dataset-mit}, with the highest available resolution of 512 pixels per degree. The motion of celestial bodies—including the Moon, Sun, and background stars—is accurately modelled using SPICE kernels obtained from the PDS Archived SPICE Data Sets~\cite{naif_pds_archived_spice}. The CRESENT-365 dataset provides near-global coverage of the lunar surface (see Fig.~\ref{fig:map_coverage}), enabling a comprehensive evaluation of our CBN pipeline. Detailed experimental results and analysis are provided in Sec.~\ref{sec:results}.

\subsection{Crater catalogue}\label{sec:crater_catalogue}

To train our CDA and evaluate the performance of our CBN pipeline, the ground truth crater correspondences of the CRESENT+ and CRESENT-365 datasets are required.   As both of these datasets contain images of the real lunar surface, a lunar crater catalogue can be used to obtain these correspondences.  

Robbins' crater catalogue is one of the most extensive lunar crater catalogues to date, detailing over 1 million craters~\cite{Robbins2019-aw}. Each crater in the catalogue is parameterised by its composition and location. Provided an image and a corresponding ground truth pose, craters from the catalogue can be projected onto the image plane using~\eqref{eq:conic_projection_expanded}, yielding ground truth imaged craters.  While Robbins' catalogue is extensive, it is not complete, meaning that every crater from an image of the lunar surface will not necessarily have a corresponding crater in the catalogue, which may affect the CID component of the pipeline.

\begin{figure}[!hbt]
    \centering
     \begin{subfigure}[b]{0.44\textwidth}
         \centering
         \includegraphics[width=\textwidth]{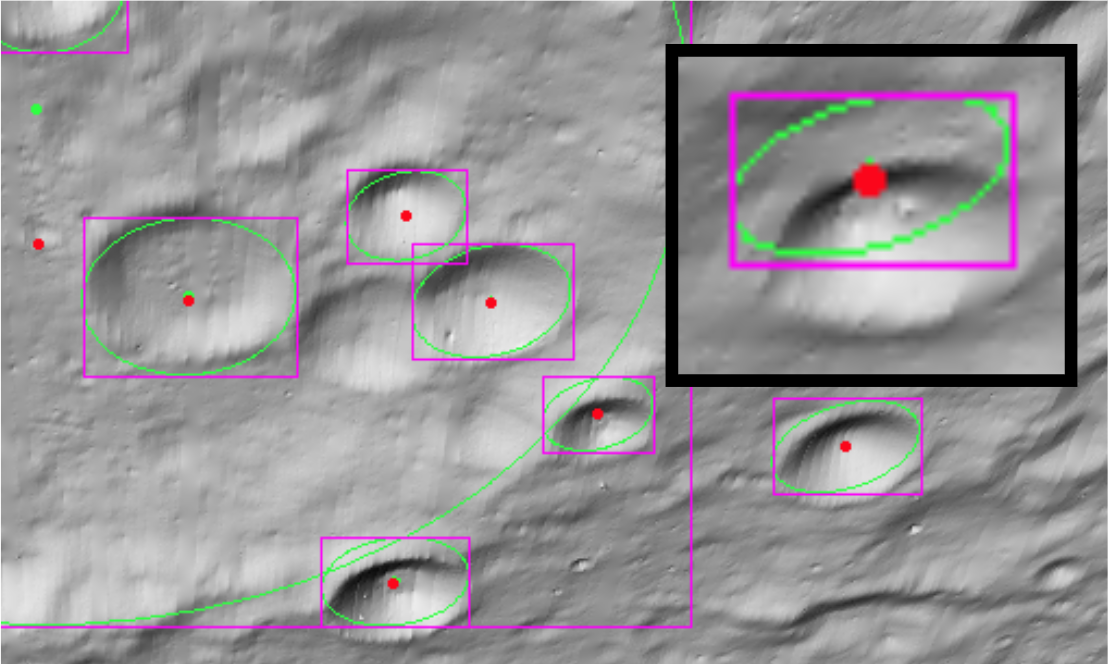}
         \caption{Before adjustment}
         \label{fig:before_rim_height_adjustment}
     \end{subfigure}
     \begin{subfigure}[b]{0.44\textwidth}
         \centering
         \includegraphics[width=\textwidth]{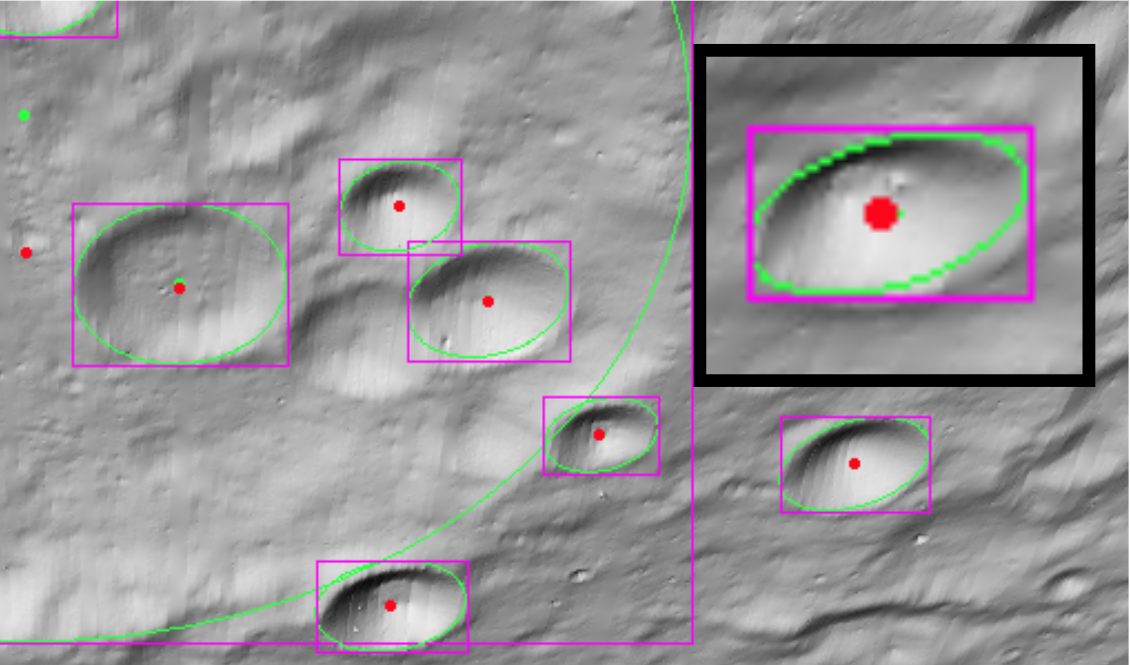}
         \caption{After adjustment}
         \label{fig:after_rim_height_adjustment}
     \end{subfigure}
    \caption{(a) Craters from Robbins' crater catalogue~\cite{Robbins2019-aw} projected on an instance of the CRESENT+ dataset~\cite{McLeod2025} do not align well with the imaged crater rims as the catalogued craters assume all craters lie on a spherical Moon. (b) Projected crater rims better align with the imaged craters when the median rim height of the crater is accounted for.}
    \label{fig:rim_height_adjustment}
\end{figure}

Another drawback of this catalogue is that the crater rim height is not provided. Without this information, crater discs are projected assuming a perfectly spherical lunar model, which causes misalignment of the projected crater rims with the true crater rims (see Fig.~\ref{fig:before_rim_height_adjustment}). This misalignment will affect all components of the CBN pipeline, including the CDA training process. For this reason, we modify the crater catalogue by adjusting the centre location of each crater to account for the median rim height of the crater.  This information was obtained by querying the corresponding lunar DEM at each crater location.  This correction process enabled many crater rims to align well with the true imaged crater rim locations, contributing to a higher quality ground truth correspondence (see Fig.~\ref{fig:after_rim_height_adjustment}).\\

In the following section, we propose our CBN pipeline for lunar mapping missions, which will be trained and evaluated on the datasets discussed in this section.

\section{The proposed pipeline}\label{sec:stella}

\begin{figure}[H]
     \centering
     \includegraphics[width=0.9\linewidth]{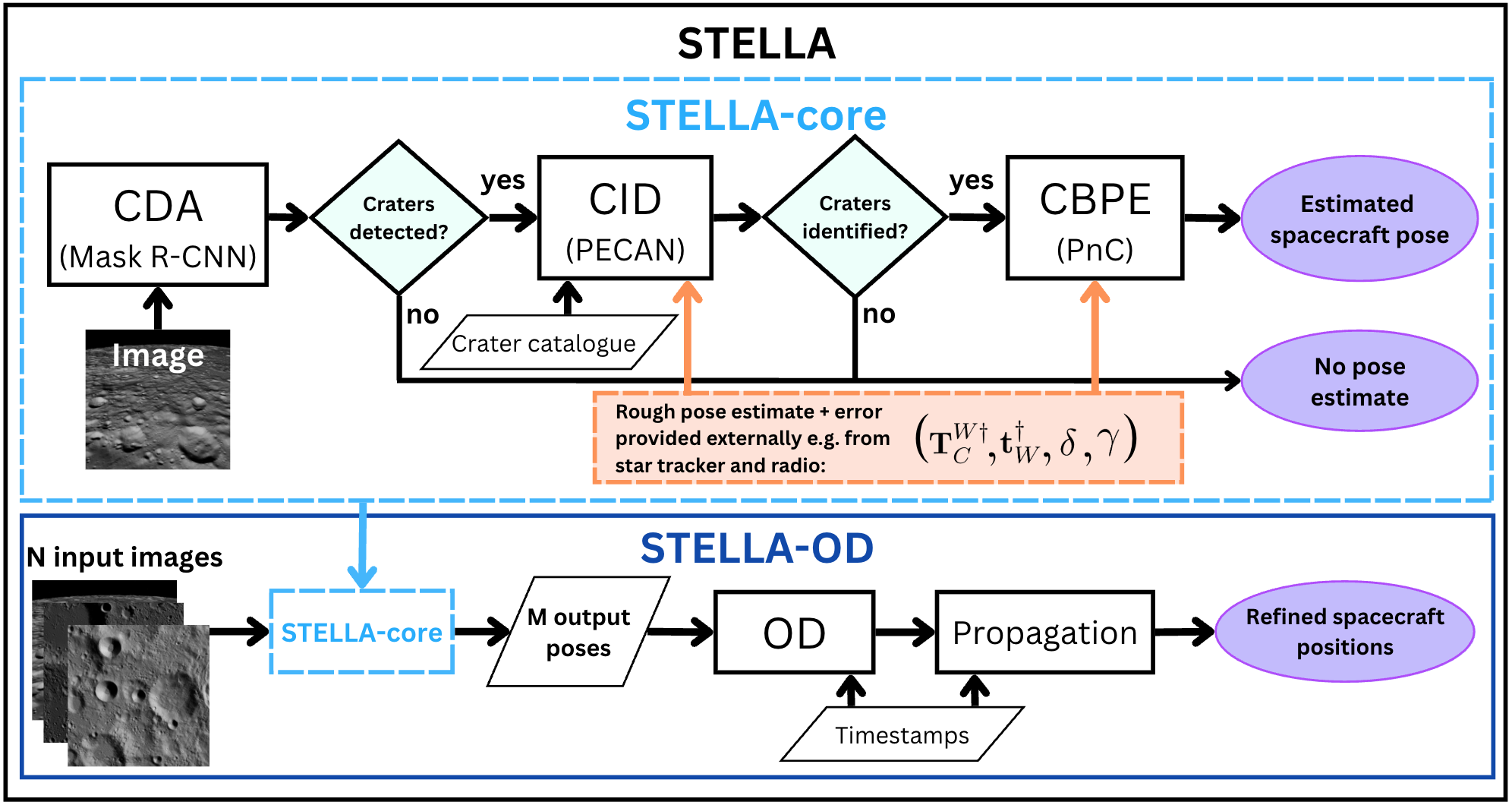}
     \caption{The proposed CBN pipeline, STELLA. STELLA-core serves as the \textit{front-end}, processing each input image to estimate the corresponding camera pose. The \textit{back-end}, STELLA-OD, refines the position estimates by fitting a single orbit to the set of position estimates by STELLA-core.}\label{fig:stella_pipeline}
\end{figure}

This work proposes a CBN pipeline, STELLA, targeting lunar mapping mission applications.  The two main components of the pipeline are STELLA-core, which performs CDA, CID and CBPE sequentially, estimating a pose per acquired image, and STELLA-OD, which performs OD on a batch of position estimates made by STELLA-core.  In this section, we detail the components of the STELLA pipeline, covering STELLA-core in Sec.~\ref{sec:stella-core} and STELLA-OD in Sec.~\ref{sec:stella-od}. A depiction of this pipeline can be seen in Fig.~\ref{fig:stella_pipeline}.

\subsection{STELLA-core}\label{sec:stella-core}
In this section, we review the CDA, CID and CBPE components of STELLA-core under various lighting conditions, global surface coverage, and non-nadir viewing angles.

\subsubsection{Crater detection algorithm}\label{sec:cda}
In this section, we discuss the architecture, training and evaluation process of the CDA component.\\

\textbf{Mask R-CNN architecture.} The CDA employed in our proposed pipeline is based on the Mask R-CNN framework~\cite{Dollar2017-qi}, employing MobileNetV2~\cite{Sandler_2018} as its backbone architecture. Mask R-CNN extends Faster R-CNN~\cite{Ren2015-fz} by concurrently predicting object bounding boxes and segmentation masks. Each predicted mask is a binary map matching the resolution of the input image, where pixels indicate the presence or absence of a crater.

\begin{figure}[!hbt]
    \centering
    \includegraphics[width=0.9\linewidth]{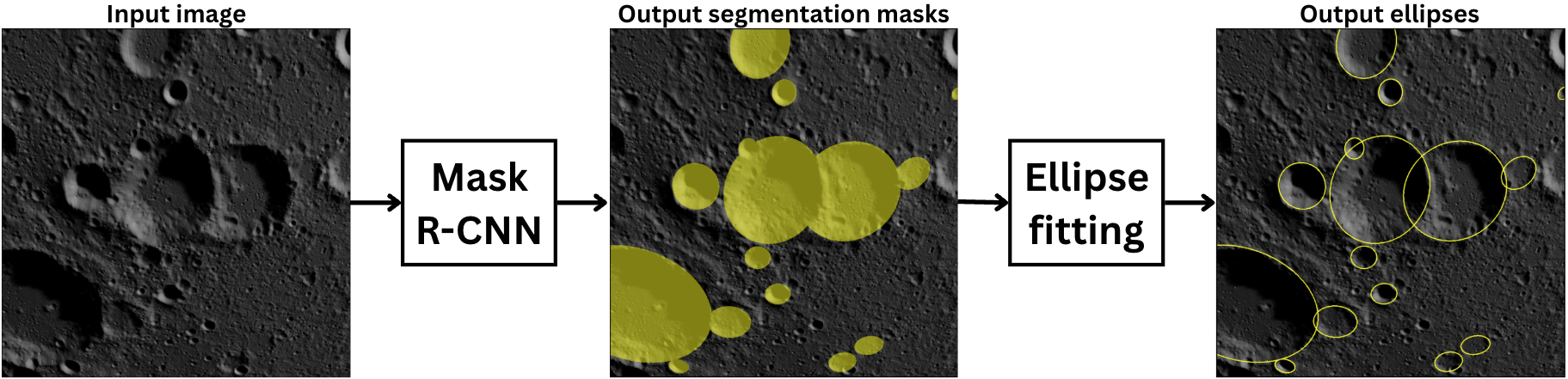}
    \caption{Mask R-CNN and ellipse fitting output from the CDA.}
    \label{fig:mask_rcnn_process}
\end{figure}

Following the approach in~\cite{Rodda2024-oz}, the segmentation masks generated by Mask R-CNN are subsequently processed using standard contour detection and ellipse-fitting methods~\cite{opencv_library}. This fitting step converts the binary masks into ellipses, which represent craters more compactly for downstream tasks. The outputs from both Mask R-CNN and the ellipse-fitting stage are illustrated in Fig.~\ref{fig:mask_rcnn_process}.

\textbf{Training.} The Mask R-CNN network was trained using only the CRESENT+ dataset~\cite{McLeod2025}, described previously in Sec.~\ref{sec:cresent}. Ground truth annotations for crater segmentation were generated by projecting (with ground truth camera poses) craters from Robbins' crater catalogue~\cite{Robbins2019-aw} onto each dataset image, as detailed in Sec.~\ref{sec:crater_catalogue}. The dataset was split into training, validation, and testing subsets at ratios of $60\%$, $20\%$, and $20\%$, respectively, resulting in 5,757 training images and 1,919 images each for validation and testing. As mentioned in Sec.~\ref{sec:cresent}, we trained our CDA only on CRESENT+ to evaluate its performance on unseen lunar surfaces and varied operational scenarios.

Due to GPU memory constraints, we applied filtering criteria to exclude certain projected craters from training. Specifically, craters that were poorly defined or too small after projection onto the image plane were omitted. We refer readers to Sec. 2 of the supplementary material for the full filtering criteria.

\begin{figure}[!hbt]
    \centering
     \begin{subfigure}[b]{0.24\textwidth}
         \centering
         \includegraphics[width=\textwidth]{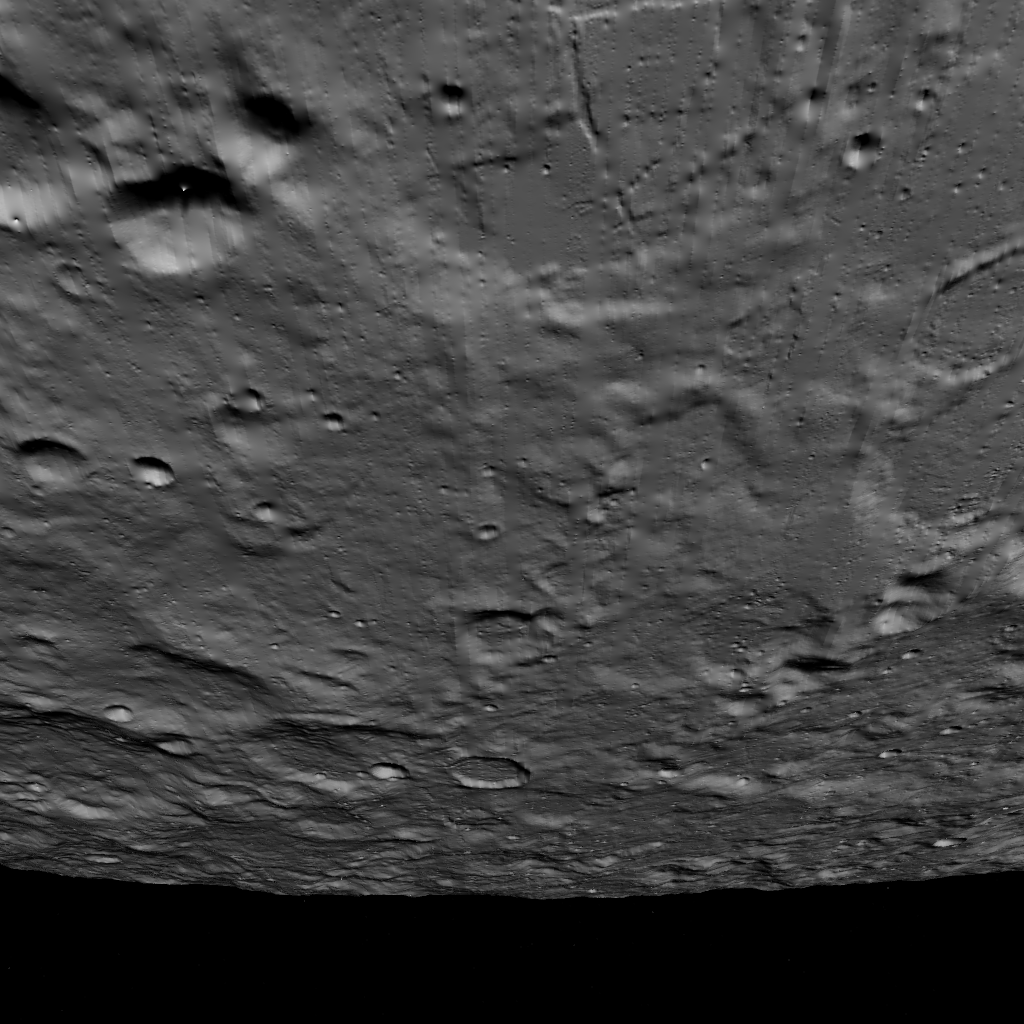}
         \caption{}
         \label{fig:original_aug}
     \end{subfigure}
     \begin{subfigure}[b]{0.24\textwidth}
         \centering
         \includegraphics[width=\textwidth]{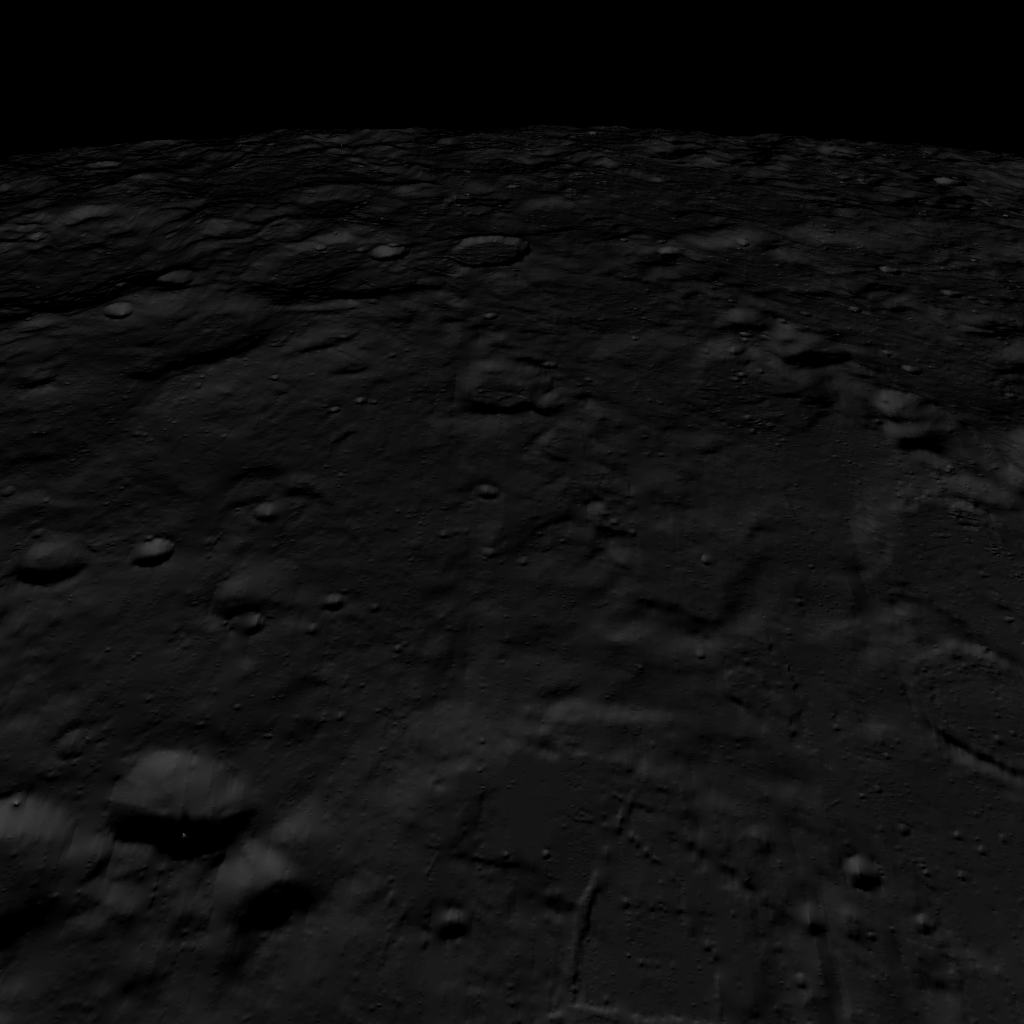}
         \caption{}
         \label{fig:dark_aug}
     \end{subfigure}
     \begin{subfigure}[b]{0.24\textwidth}
         \centering
         \includegraphics[width=\textwidth]{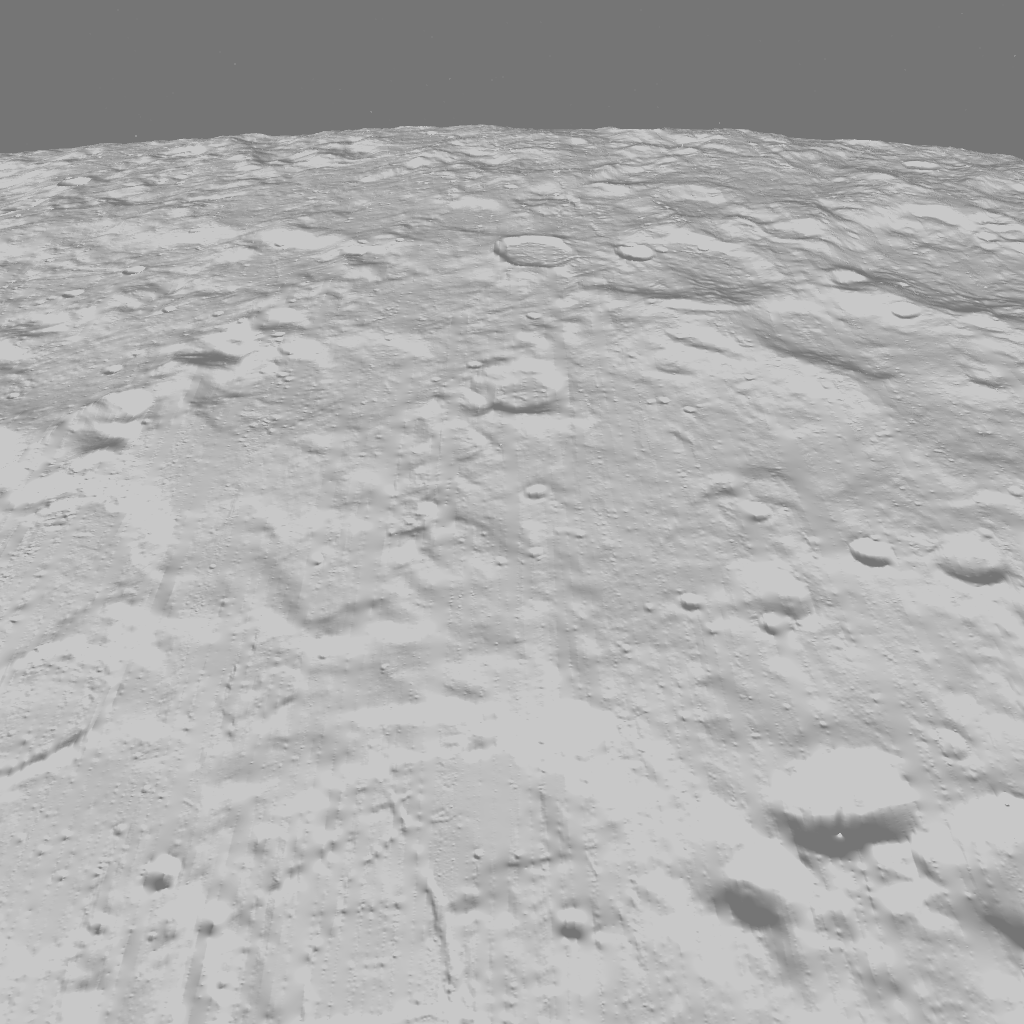}
         \caption{}
         \label{fig:low_contrast_aug}
     \end{subfigure}
     \begin{subfigure}[b]{0.24\textwidth}
         \centering
         \includegraphics[width=\textwidth]{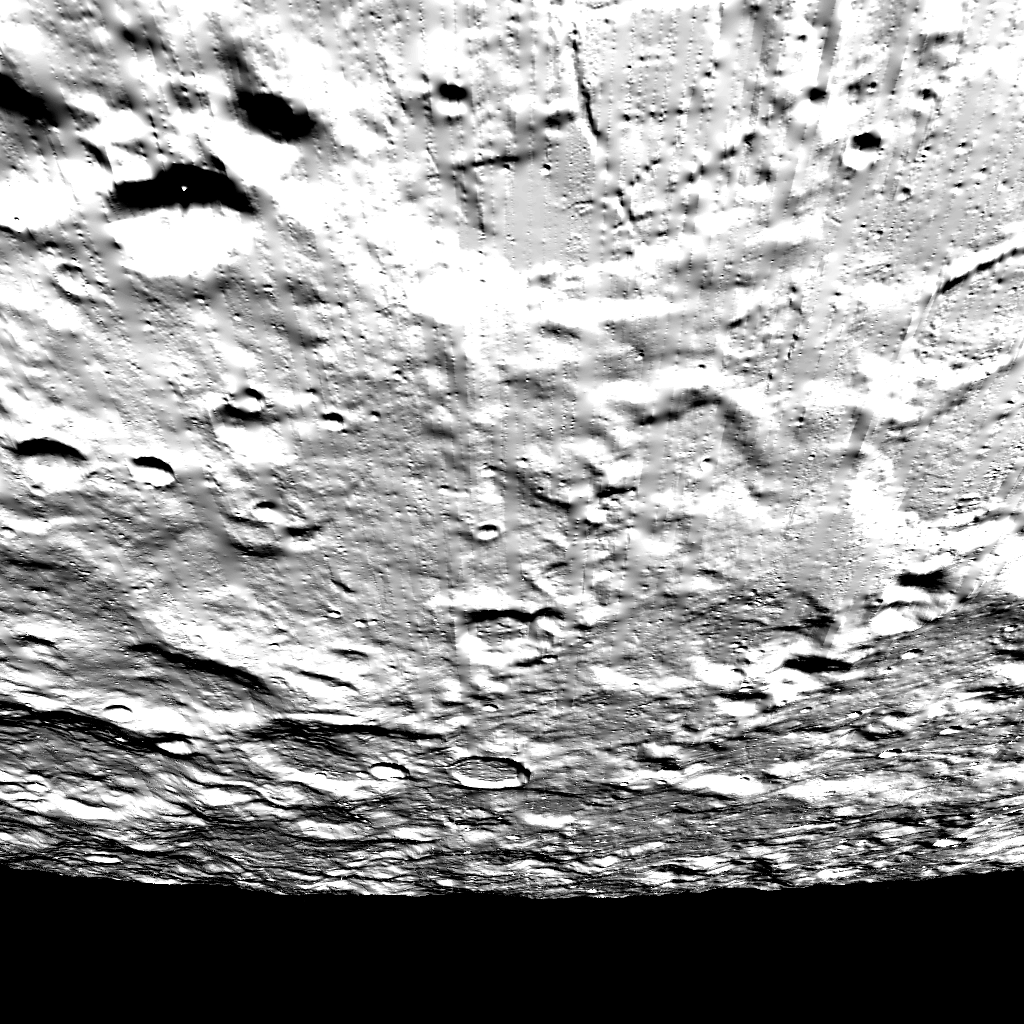}
         \caption{}
         \label{fig:high_contrast}
     \end{subfigure}
    \caption{Data augmentation examples for a single CRESENT+ instance. (a) No augmentation, (b) low brightness, vertical flip, (c) low contrast, horizontal and vertical flip, (d) high brightness, high contrast.}
    \label{fig:data_augmentation}
\end{figure}

To improve the robustness and generalisation capabilities of the network, we implemented several data augmentation techniques during training. Since the CRESENT+ dataset primarily contains well-illuminated images, brightness and contrast augmentations were applied to mimic the more diverse illumination conditions present in the CRESENT-365 dataset. Additionally, horizontal and vertical flipping augmentations were applied to images and their associated ground truth masks. Examples of these augmentation methods are shown in Fig.~\ref{fig:data_augmentation}.

The training was conducted on an NVIDIA GeForce RTX 2080 Ti GPU using the PyTorch framework. The model was trained on CRESENT+ for 43 epochs using the stochastic gradient descent (SGD) optimiser~\cite{pytorch_sgd} with a learning rate of 0.01, a momentum of 0.9, and a weight decay of 0.0005. While evaluating the CRESENT-365 set, we found that the early checkpoints (15th epoch) performed better, suggesting a distribution gap between the two datasets. 


\textbf{Evaluation.}  The CDA was evaluated on the test set of CRESENT+ and the CRESENT-365 dataset. Crater detections made by the CDA returned a confidence value $\hat{c}$ per prediction, where the prediction was retained if $\hat{c} \geq \bar{c}$. The confidence threshold $\bar{c}$ was tuned to obtain the highest F1 score~\cite{Chicco2020-nq}. A crater detection was considered a true positive if the largest intersection-over-union (IoU) with a ground truth crater mask was greater than 50\%. We found the confidence thresholds that achieved the highest F1 score for the CRESENT+ test set and CRESENT-365 dataset were 0.6 and 0.5 respectively.

All retained crater detections made by the CDA were returned as imaged ellipses for input to the following CID component of the STELLA-core pipeline. For data instances where no crater detections were retained, STELLA-core terminates and returns `no-result'.

\subsubsection{Crater identification}\label{sec:cid}
We employed PECAN as the CID component of STELLA due to its flexibility and robust performance under general scenarios. Specifically, PECAN does not impose constraints on the camera pointing angle, terrain geometry, or crater shape, making it particularly suitable for lunar mapping missions. 

PECAN operates on the ellipse representation for the set of detected craters, denoted as $\mathsf{D} = \{\mathcal{D}_i\}^I_i$, and the conic representation for the set of catalogued craters $\mathsf{C}=\{\mathbf{C}_j\}^J_j$. The method assumes prior knowledge of the intrinsic calibration matrix\footnote{The camera can be calibrated pre- or post-launch~\cite{de2014automatic, hartley1999camera}.} $\mathbf{K}$ and the attitude ${\mathbf{T}^{W}_{C}}^{\dagger}$ of the camera. Given these fixed parameters, PECAN seeks to estimate the optimal camera position that maximises the number of matches between detected craters ($\mathsf{D}$) and catalogued craters ($\mathsf{C}$). Formally, this optimisation is defined as follows,
\begin{align}\label{eq:PECAN_opt}
\begin{aligned}
& \underset{\mathbf{t}_{W} \in \mathbb{R}^3}{\text{max}}
& & m(\mathbf{t}_{W} \mid \mathsf{D}, \mathsf{C},{\mathbf{T}^W_C}^\dagger,\mathbf{K}).
\end{aligned}
\end{align}
The objective function, $m$, in \eqref{eq:PECAN_opt} quantifies the number of crater matches between sets $\mathsf{D}$ and $\mathsf{C}$, and is explicitly given by:

\begin{equation}\label{eqn:PECAN_obj}
    m(\mathbf{t}_{W} \mid \mathsf{D}, \mathsf{C}, {\mathbf{T}^{W}_{C}}^{\dagger},\mathbf{K}) = \sum_{i=1}^I \max_{j=1,\dots,J} \mathbb{I}\left( f(\mathcal{D}_i, \mathbf{C}_j \mid \mathbf{K},{\mathbf{T}^{W}_{C}}^{\dagger},\mathbf{t}_{W}) \le \epsilon \right) .
\end{equation}
Here, the indicator function $\mathbb{I}$ returns `1' if the metric function $f$ yields a value less than a threshold $\epsilon$, which we set to 20 following~\cite{chng2024crater}. The $\max$ operation ensures each detected crater is matched with at most one catalogued crater. Intuitively, craters are considered matched when their projected positions are sufficiently close. 
PECAN employs the \textit{ellipse distance} as its metric $f$, which computes an algebraic error directly comparing ellipse parameters, defined by:
\begin{equation}\label{eq:residual}
    f(\mathcal{D}_i, \mathbf{C}_j \mid \mathbf{K}, {\mathbf{T}^{W}_{C}}^{\dagger}, \mathbf{\mathbf{t}}_{W}) := \left\| \mathcal{D}_i - v(\mathbf{C}_j \mid \mathbf{K}, {\mathbf{T}^{W}_{C}}^{\dagger},\mathbf{\mathbf{t}}_{W}) \right\|_2,
\end{equation}
\noindent where the function $v$, as defined in~\eqref{eq:conic_to_ellipse}, projects the catalogued crater to IRF and extracts its ellipse representation.

The algorithm to solve~\eqref{eq:PECAN_opt} can be referred to in~\cite{chng2024crater}. In essence, PECAN invokes a Perspective-1-Ellipsoid solver~\cite{gaudilliere2023perspective} to obtain a position estimate $\mathbf{\mathbf{t}}_{W}$ for each detected-catalogued crater pair, and evaluates the quality of $\mathbf{\mathbf{t}}_{W}$ with~\eqref{eqn:PECAN_obj}. The optimal solution to~\eqref{eq:PECAN_opt} then leads to the CID output, \ie, correspondence set of detected and catalogued craters, as follows,
\begin{equation}\label{eqn:identity}
    {\mathcal{M}'} = \{ \langle i, j \rangle \mid f(\mathcal{D}_i, \mathbf{C}_j \mid \mathbf{K},{\mathbf{T}^{W}_{C}}^{\dagger},\mathbf{\mathbf{t}}'_{W}) \leq \epsilon \}.
\end{equation}
\noindent There are two exception cases to \eqref{eqn:identity}. First, a single $i$ may correspond to several candidate $j$; in which case, we retain the $j$ producing the lowest error in \eqref{eq:residual}. Second, an $i$ might have no corresponding $j$ and will not be included in the set. Consequently, the maximum achievable value of \eqref{eqn:PECAN_obj}, and equivalently, the cardinality of ${\mathcal{M}'}$ is at most $I$,~\ie, the number of detected craters, and it typically falls short due to noise and spurious crater detections present in set $\mathsf{D}$. As reported in~\cite{chng2024crater}, it is often sufficient to terminate the algorithm when a sufficient number of matches is achieved. We found the same in our evaluation: terminating PECAN when the number of matches found (with~\eqref{eq:PECAN_opt}) is more than $60\%$ of the number of detected craters, leads to reliable CID results while reducing computational time. Note that if the criterion is not matched at the end of PECAN, STELLA-core terminates and returns `no-result'. 

CID methods typically operate on a subset of well-defined and visible craters\footnote{As determined by the altitude of the mission profile.} extracted from the full crater catalogue to construct a mission-specific catalogue~\cite{Christian2021-ts, chng2024crater, hanak2009lost}. These mission catalogues are generally much smaller than the full catalogue; for instance, Hanak~\etal~\cite{hanak2010crater} utilised a mission catalogue with approximately 4,000 craters, while Christian~\etal~\cite{Christian2021-ts} proposed one with about 20,000 craters. In contrast, the full Robbins catalogue contains over one million craters. Such an extraction is conducted to improve the CID performance (by increasing the uniqueness of each descriptor) and reduce the memory consumption. However, this extraction process introduces a risk: craters detected by the CDA that are not included in the mission catalogue will be treated as outliers during CID, as no corresponding matches exist. This, in turn, reduces the number of correspondences available for the subsequent CBPE step. To mitigate this risk, we employ the full Robbins catalogue as the mission catalogue in our experiments. We highlight that this is further enabled by adopting the descriptor-less CID method, PECAN. Unlike traditional descriptor-based CID methods, which exhibit exponential memory growth even with optimised expansion strategies~\cite{Christian2021-ts}, PECAN scales linearly with the size of the mission catalogue, making it well-suited for the adoption of large-scale mission catalogues.

While employing the full Robbins catalogue avoids the \textit{missing crater} problem, it also enlarges the search space, potentially slowing CID and increasing the likelihood of spurious matches. To keep the tractability of crater matching without sacrificing completeness, we exploit the coarse pose information that is routinely available on most lunar missions: a position estimate from radio-ranging with uncertainty of $\sim6.7$ km and an attitude estimate from a star tracker with uncertainty of $\sim0.01^\circ$~\cite{McLeod2024-ui,2000-rq}. To stress-test the robustness of our pipeline, we adopt more conservative uncertainties, \ie, 11 km in position and $0.02^\circ$ in attitude. Given this uncertain pose, we extract at run time a visibility sub-catalogue: the subset of craters that could lie within the camera’s field of view. This on-the-fly pruning keeps the computational load manageable and focuses PECAN on the most relevant portion of the large catalogue, while still guaranteeing that any crater detected by the CDA has a high probability of having a counterpart in the mission catalogue.

\subsubsection{Crater-based pose estimation}\label{sec:cbpe}

Once the 3D–2D correspondences between catalogued and detected craters are established, STELLA-core estimates the full 6DoF camera pose using its CBPE component. For this task, we adopt the perspective-n-crater (PnC) algorithm~\cite{McLeod2024-ui}, which has been identified through extensive evaluation in prior work as one of the most suitable CBPE methods for CBN~\cite{McLeod2025, McLeod2024-ui}.

Given the correspondence set from CID, ${\mathcal{M}'}$, the PnC problem is formulated as,
\begin{equation}\label{eq:pnc_masked}
\begin{aligned}
\min_{(\mathbf{T}^{W}_{C},\,\mathbf{t}_{W})\in SE(3)}
      &\;
      \sum_{\langle i,j\rangle\in\mathcal{M}'}
      \chi(f(\mathcal{D}_{i},\;
               v(\mathbf{T}^{W}_{C},\mathbf{t}_{W}
                        \mid \mathbf{C}_{j},\mathbf{K}))), \\[4pt]
\text{subject to}\quad
      &\angle\!\bigl(\mathbf{T}^{W}_{C},\,{\mathbf{T}^{W}_{C}}^{\dagger}\bigr)\;\le\;\delta, \\[2pt]
      &\bigl\|\,\mathbf{t}_{W}-\mathbf{t}_{W}^{\dagger}\bigr\|_{2}\;\le\;\gamma .
\end{aligned}
\end{equation}

\begin{figure}[!hbt]
    \centering
    \includegraphics[width=0.5\linewidth]{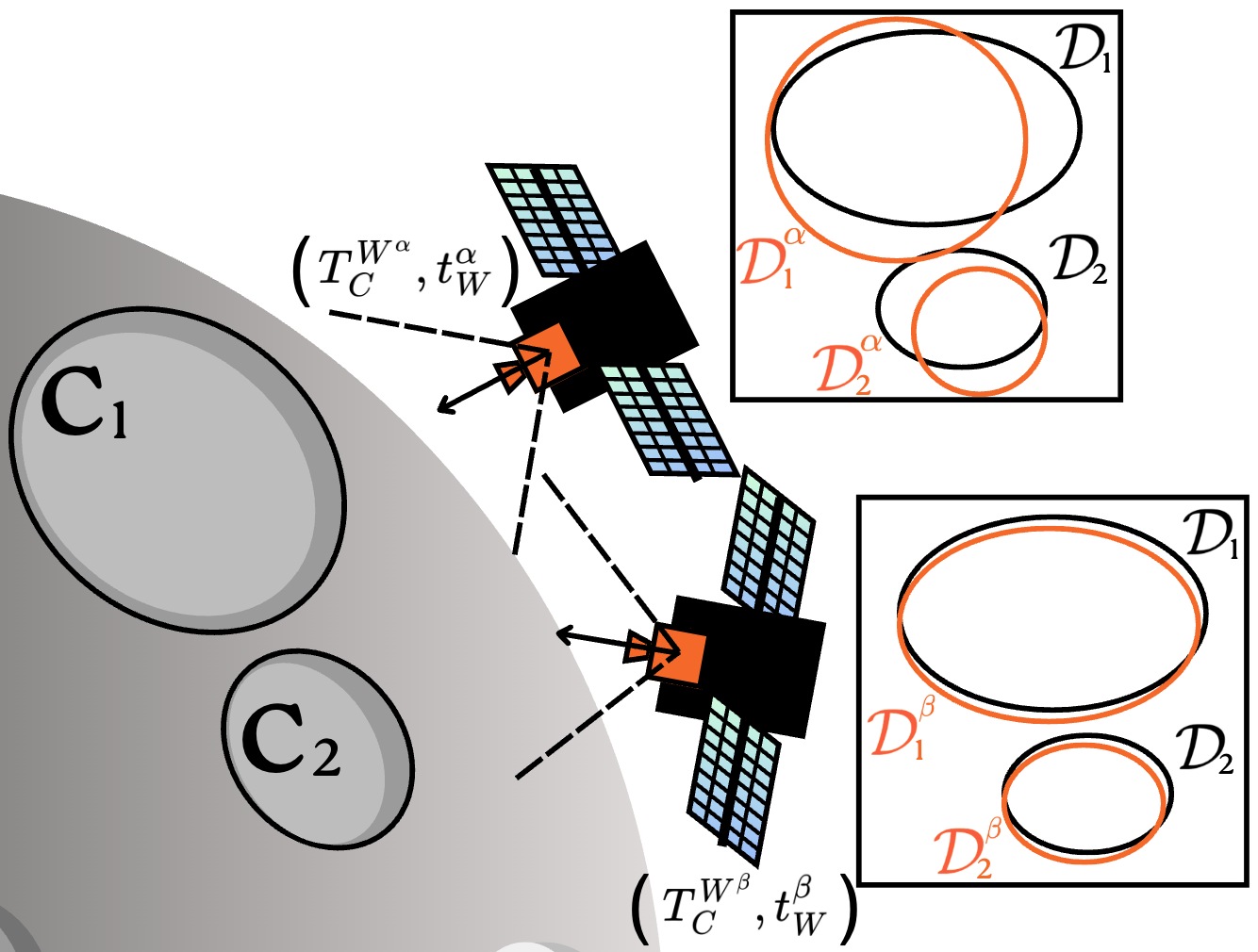}
    \caption{The PnC optimisation framework minimises the reprojection error of the detected crater rims (black ellipses) with the reprojected rims (orange ellipses). In this example, the optimised pose $(\mathbf{T}^{W^\beta}_C,\mathbf{t}^\beta_W)$ better aligns the detected craters with the reprojected craters from the initialised pose $(\mathbf{T}^{W^\alpha}_C,\mathbf{t}^\alpha_W)$.}
    \label{fig:pnc_diagram}
\end{figure}

In essence,~\eqref{eq:pnc_masked} seeks the optimal pose that minimises the ellipse distance, \ie, $f$ as defined in~\eqref{eq:residual}, between the projected catalogued craters and their corresponding match in the detected crater set. To mitigate the effect of potential false correspondences in the output of CID, PnC adopts an M-estimator framework. Specifically, it employs Tukey’s biweight loss function~\cite{Zhang1997-in}, denoted by $\chi$, which downweights outliers and is defined as follows:

\begin{equation}\label{eq:tukey}
    \chi(f) = 
    \begin{cases}
        (\alpha^2/6)(1-[1-(f/\alpha)^2]^3) &\text{ if } |f| \leq \alpha\\
        (\alpha^2/6) & \text{otherwise}
    \end{cases} ,
\end{equation}

\noindent where $\alpha$ is a positive, tunable hyperparameter that defines the influence range of each correspondence. Matches with ellipse distances $|f| \leq \alpha$ contribute to the optimisation, while those exceeding this threshold are treated as outliers and its error is trimmed fixed at $(\alpha^2/6)$. For further M-estimation implementation details, we refer the reader to~\cite{McLeod2024-ui}. The key elements of PnC are visualised in Fig.~\ref{fig:pnc_diagram}.

PnC assumes a priori known pose $({\mathbf{T}^W_C}^\dagger,\mathbf{t}_W^\dagger)$ is known with uncertainty $\delta$ and $\gamma$ respectively. The attitude uncertainty, $\delta$, is set to $0.02^\circ$ and the position uncertainty, $\gamma$, is set to 11km (see Sec.~\ref{sec:cid}). PnC was implemented using SciPy's Optimize with L-BGFS-B method and M-estimators was employed through an iteratively reweighted least squares algorithm~\cite{McLeod2024-ui}. \\

\subsection{STELLA-OD}\label{sec:stella-od}
In lunar mapping missions, the trajectory of the spacecraft typically follows an orbit, which makes it advantageous to incorporate an orbit determination (OD) component into a CBN pipeline. The effectiveness of incorporating OD has been demonstrated in missions such as NASA’s Lunar Reconnaissance Orbiter (LRO)\cite{slojkowski2015orbit} and ISRO’s Chandrayaan-1\cite{vighnesam2009precise}.

By constraining the estimated spacecraft positions to obey an underlying orbital model, this component can help correct \textit{outlying} position estimates\footnote{Position estimates that deviate far from the least-squares orbit estimate.} produced by the STELLA-core. Moreover, the estimated orbit enables forecasting of future spacecraft positions, which can serve either as an additional initialisation source for subsequent CBN runs or as a fallback solution in frames where CBN yields `\textit{no result}' as a result of extreme lighting conditions or in regions of the lunar surface with sparse craters (see Sec.~\ref{sec:cresent-365_results}, which highlights the degraded performance of STELLA in such scenarios).

\subsubsection{Orbit determination}\label{Sec:OD}

Given $M$ position estimates obtained from CBPE, which we denote as $\{\mathbf{t}^{(\mathrm{m})}_{W}\}^M_{\mathrm{m}=1}$, the OD task is formulated as a minimisation problem below, 
\begin{eqnarray}\label{eqn:OD} \underset{\mathbf{o} \in \mathbb{R}^6}{\text{min}} \ \frac{1}{M} \sum_{\mathrm{m}=1}^{M} \ \parallel p(\mathbf{o}; t_\mathrm{m}) - \mathbf{t}^{(\mathrm{m})}_{W} \parallel_2 \ \ .\end{eqnarray} 
\noindent where $\mathbf{o} \in \mathbb{R}^6$ is the Keplerian orbital elements (KOE), and $p$ represents a function that converts the KOE into an initial state vector and propagates it to a specific timestamp $t_\mathrm{m}$ using a Keplerian two-body propagator~\cite[Chap.2]{vallado2001fundamentals}. Intuitively, the goal is to find the optimal orbital parameters that minimise the discrepancy between the propagated positions and the position estimates obtained from CBPE. We solve~\eqref{eqn:OD} using the L-BFGS-B algorithm implemented in SciPy, consistent with the approach used in our CBPE solver. 

As depicted in the bottom half of Fig.~\ref{fig:stella_pipeline}, the optimal orbit from solving~\eqref{eqn:OD} can be propagated with function $p$ to all the timestamps associated with the input images to obtain a set of \textit{refined} position estimates. 

To initialise \eqref{eqn:OD}, we first need to determine a velocity vector\footnote{The state vector representation of an orbit comprises a position vector and a velocity vector.} from the set of position estimates $\{\mathbf{t}^{(\mathrm{m})}_{W}\}^M_{\mathrm{m}=1}$. For this purpose, we applied Gibbs method~\cite[Chap.7]{vallado2001fundamentals}, which estimates the spacecraft’s velocity vector from three known position vectors.\\

In this section, we described the CDA, CID, and CBPE components that form the STELLA-core pipeline, which estimates the spacecraft pose from a single image. We also introduced the OD component in the STELLA-OD pipeline, which estimates the orbital trajectory based on a series of pose estimates produced by STELLA-core. The overall structure of STELLA-core and STELLA-OD is illustrated in Fig.~\ref{fig:stella_pipeline}, and their performance is evaluated in the following section.

\section{Results}\label{sec:results}

In this section, we evaluate the performance of our proposed STELLA pipeline in the context of lunar mapping missions. 

We begin by describing the pose initialisation process (Sec.~\ref{sec:pose_init}), followed by the evaluation metrics (Sec.~\ref{sec:evaluation_metrics}), and then the data attributes of the CRESENT-365 dataset (Sec.~\ref{sec:cresent-365_illumination_conditions}).

While the primary focus of this work is to evaluate the overall performance of the STELLA pipeline, we first analyse the performance of our CDA in isolation from the rest of the pipeline (Sec.~\ref{sec:analysis_of_cda}).  This is the only component of STELLA-core that depends on the visual qualities of the input image, providing insights into the visual challenges of the benchmark datasets.  As the CID component relies on the quality of crater detections made by the CDA, which was studied extensively by previous CID works~\cite{Christian2021-ts, chng2024crater}, we do not separately analyse its performance in this work.

Finally, as the central focus of this work, we evaluate STELLA under the CRESENT+ test set and CRESENT-365 dataset, providing insights into its performance under the expected conditions of lunar mapping missions (Sec.~\ref{sec:results_on_cresent} -~\ref{sec:cresent-365_results}).

\subsection{Pose initialisation}\label{sec:pose_init}

The ground truth pose ($\mathbf{T}^{W\star}_C, \mathbf{t}^\star_W$) is provided for each data instance of the CRESENT+ and CRESENT-365 datasets.  Therefore, a corresponding pose estimate (${\mathbf{T}^{W}_C}^\dagger, \mathbf{t}^\dagger_W$) was randomly initialised per data instance, where, $\mathbf{T}^{W\star}_C - \delta \leq {\mathbf{T}^{W}_C}^\dagger \leq \mathbf{T}^{W\star}_C + \delta$ and $\mathbf{t}^\star_W - \gamma \leq \mathbf{t}^\dagger_W \leq \mathbf{t}^\star_W + \gamma$.

This pose is first passed to CID, which uses both the pose estimate and error uncertainty to restrict the crater catalogue and identify craters. The CID returns an optimised position estimate, which needs to be further refined by CBPE. 

CBPE is initialised with $({\mathbf{T}^{W}_C}^\dagger, \mathbf{t}^\dagger_W)$ where $\mathbf{t}^\dagger_W$ is the position estimate returned by CID if it lies within an error $\gamma$ of the original initialised position.

\subsection{Evaluation metrics}\label{sec:evaluation_metrics}

In this work, we consider the following evaluation metrics for CDA performance: \textbf{precision}, which measures the quality of crater detections; \textbf{recall}, which measures the quantity of crater detections; and \textbf{F1 score}, which is the harmonic mean of precision and recall~\cite{Chicco2020-nq}.

The entire STELLA pipeline will be assessed on pose accuracy and corresponding surface localisation accuracy. Pose accuracy considers the following metrics from~\cite{McLeod2024-ui,McLeod2025}:

\begin{itemize}
    \item \textbf{position error (m):} $|\mathbf{t}_W-\mathbf{t}^*_W|$ 
    \item \textbf{angular error (deg):} $\arccos\left(\dfrac{\text{Tr}(\mathbf{T}^W_C\cdot\mathbf{T}^{W*^T}_C)-1}{2}\right)$ 
    \item \textbf{observed surface error (m):} $|\mathbf{m}_W-\mathbf{m}^*_W|$ - the straight line distance between the point of intersection of the line of sight of the camera with the lunar surface under estimated and ground truth camera poses (points $\mathbf{m}_W$, $\mathbf{m}^*_W$ respectively).
\end{itemize}

\subsection{CRESENT-365 data attributes}\label{sec:cresent-365_illumination_conditions}

The data instances from CRESENT-365 contain a number of attributes that will affect the performance of STELLA. 

\begin{figure}[!hbt]
    \centering
    \includegraphics[width=0.6\linewidth]{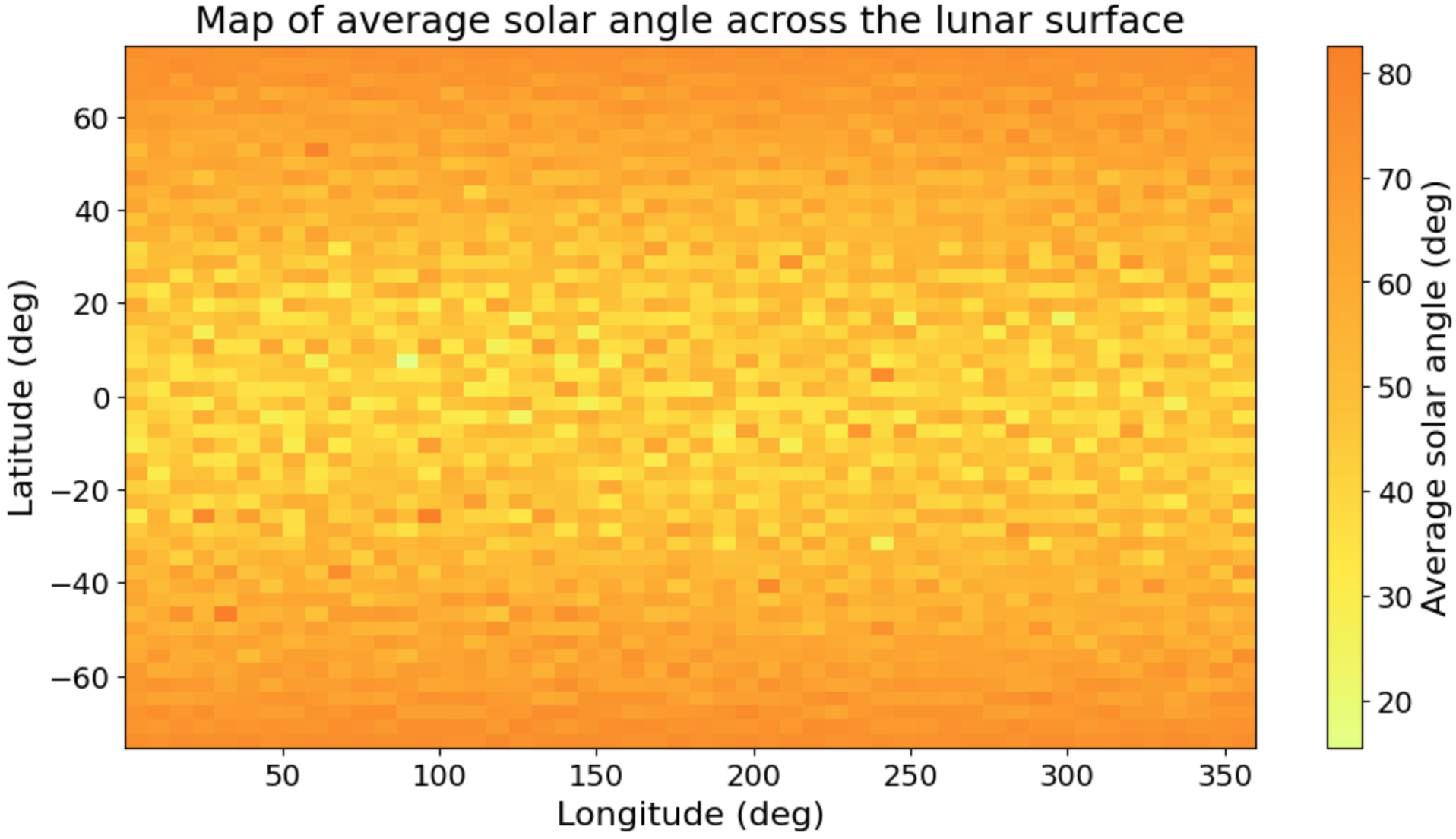}
    \caption{Average solar angle of CRESENT-365 visualised across the lunar surface. The surface with a lower solar angle value is brighter.}
    \label{fig:map_solar_orbital}
\end{figure}

One such attribute that will later be evaluated in isolation is solar angle~\eqref{eqn:solar_angle},~\ie, how illuminated the observed surface appears. Fig.~\ref{fig:map_solar_orbital} details the average solar angle of CRESENT-365, visualised across a map of the lunar surface (see Fig.~\ref{fig:map_coverage} for the corresponding map of data instance frequency). From this figure, we observe brighter illumination at $0^\circ$ latitude, becoming darker near the North and South Poles.

\begin{figure}[!hbt]
    \centering
     \begin{subfigure}[b]{0.65\textwidth}
        \centering
        \includegraphics[width=\textwidth]{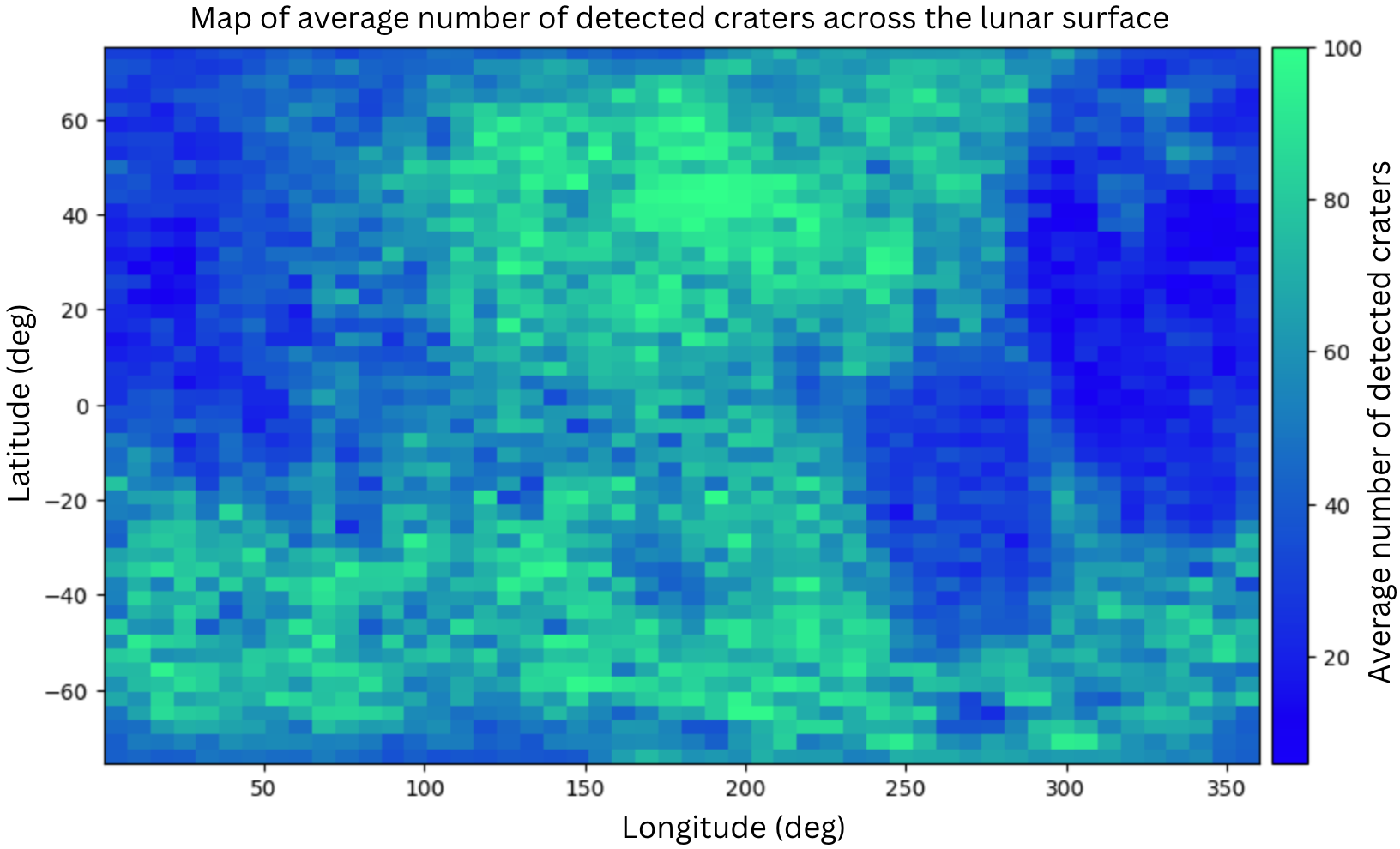}
        \caption{}
        \label{fig:number_of_detected_craters}
     \end{subfigure}
     \begin{subfigure}[b]{0.85\textwidth}
        \centering
        \includegraphics[width=\textwidth]{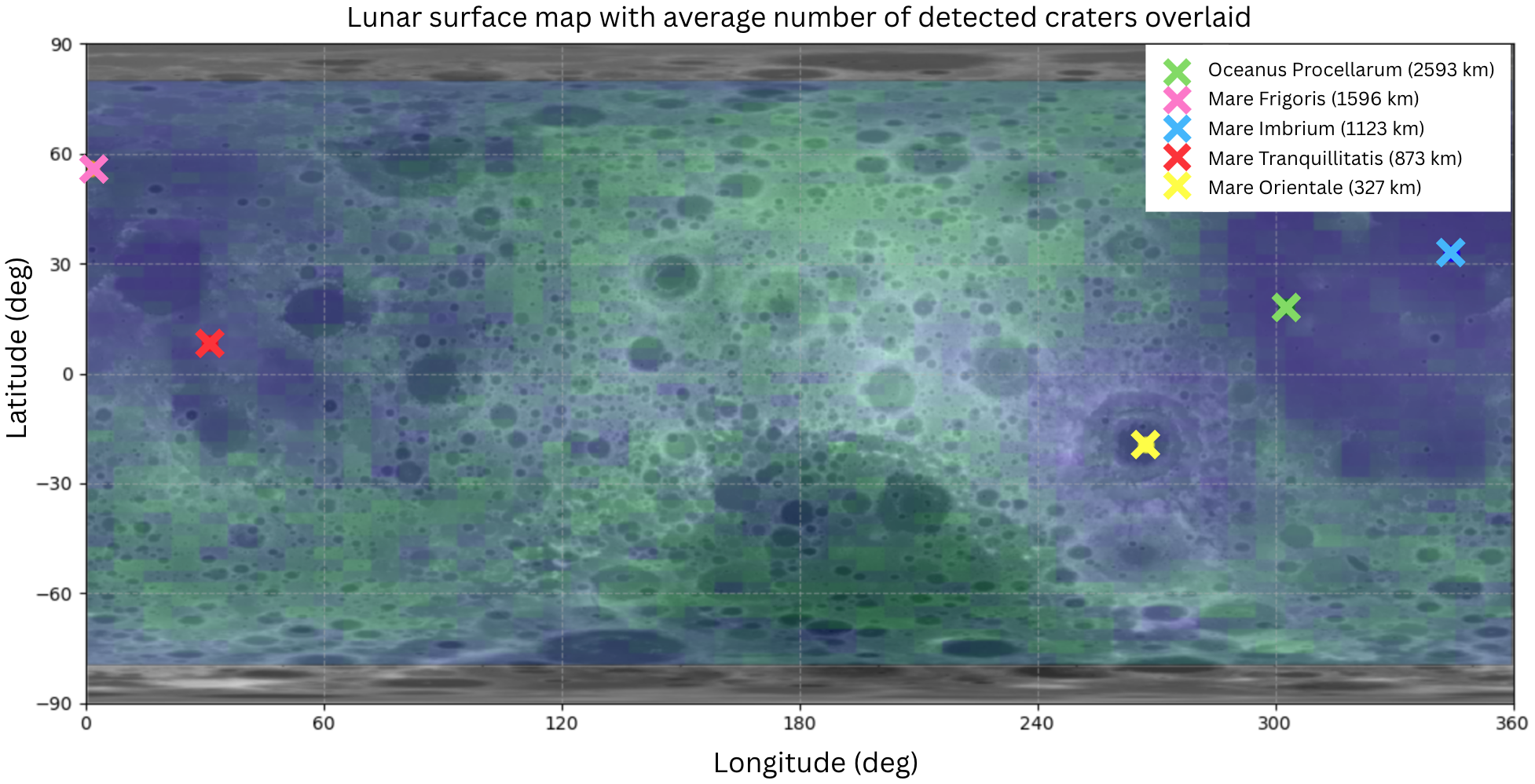}
        \caption{}
        \label{fig:map_overlay}
     \end{subfigure}
     \label{}
     \caption{(a) Average number of craters detected by our CDA on CRESENT-365. (b) Average number of detected craters from (a) overlaid onto a DEM image of the lunar surface, highlighting some of the notable Lunar Maria regions (points on the map are the centre of the region, with the radius of the regions specified in the legend).}
\end{figure}

\begin{figure}[!hbt]
    \centering
     \begin{subfigure}[b]{0.24\textwidth}
        \centering
        \includegraphics[width=\textwidth]{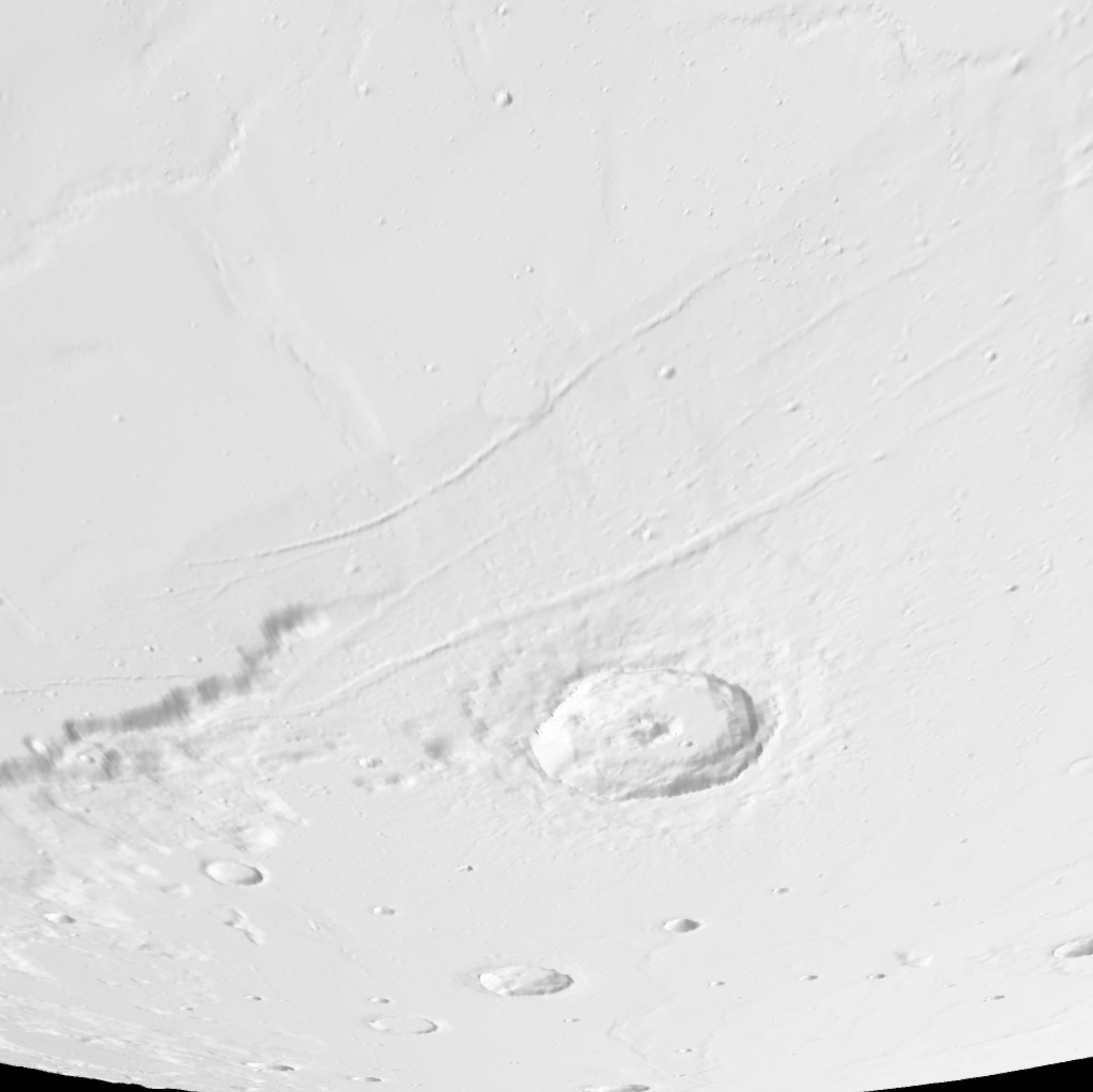}
        \caption{Mare Tranquillitatis}
        \label{fig:4370_mare_tranquillitius}
     \end{subfigure}
     \begin{subfigure}[b]{0.24\textwidth}
        \centering
        \includegraphics[width=\textwidth]{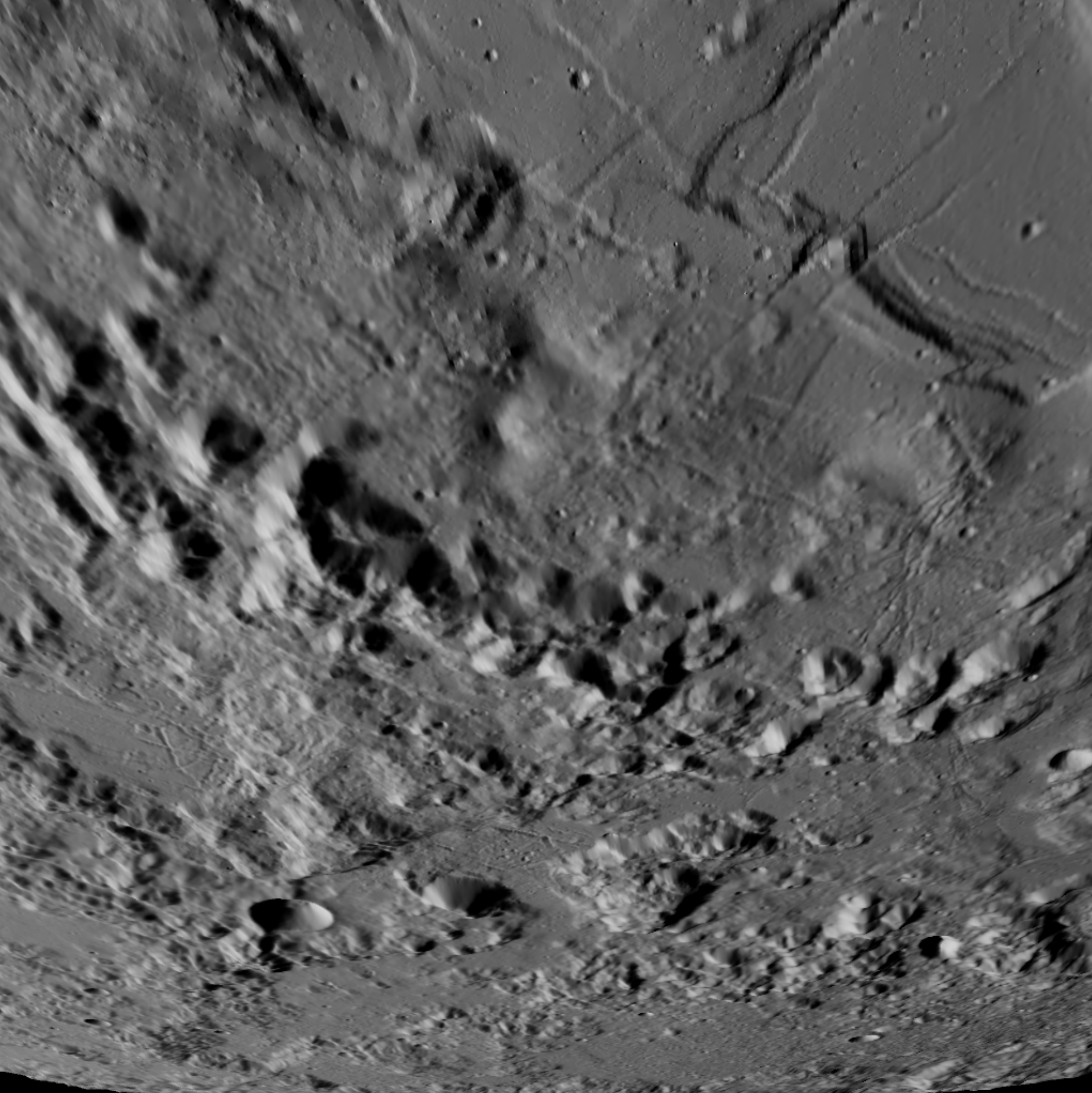}
        \caption{Mare Orientale}
        \label{fig:293_mare_orientale}
     \end{subfigure}
     \begin{subfigure}[b]{0.24\textwidth}
        \centering
        \includegraphics[width=\textwidth]{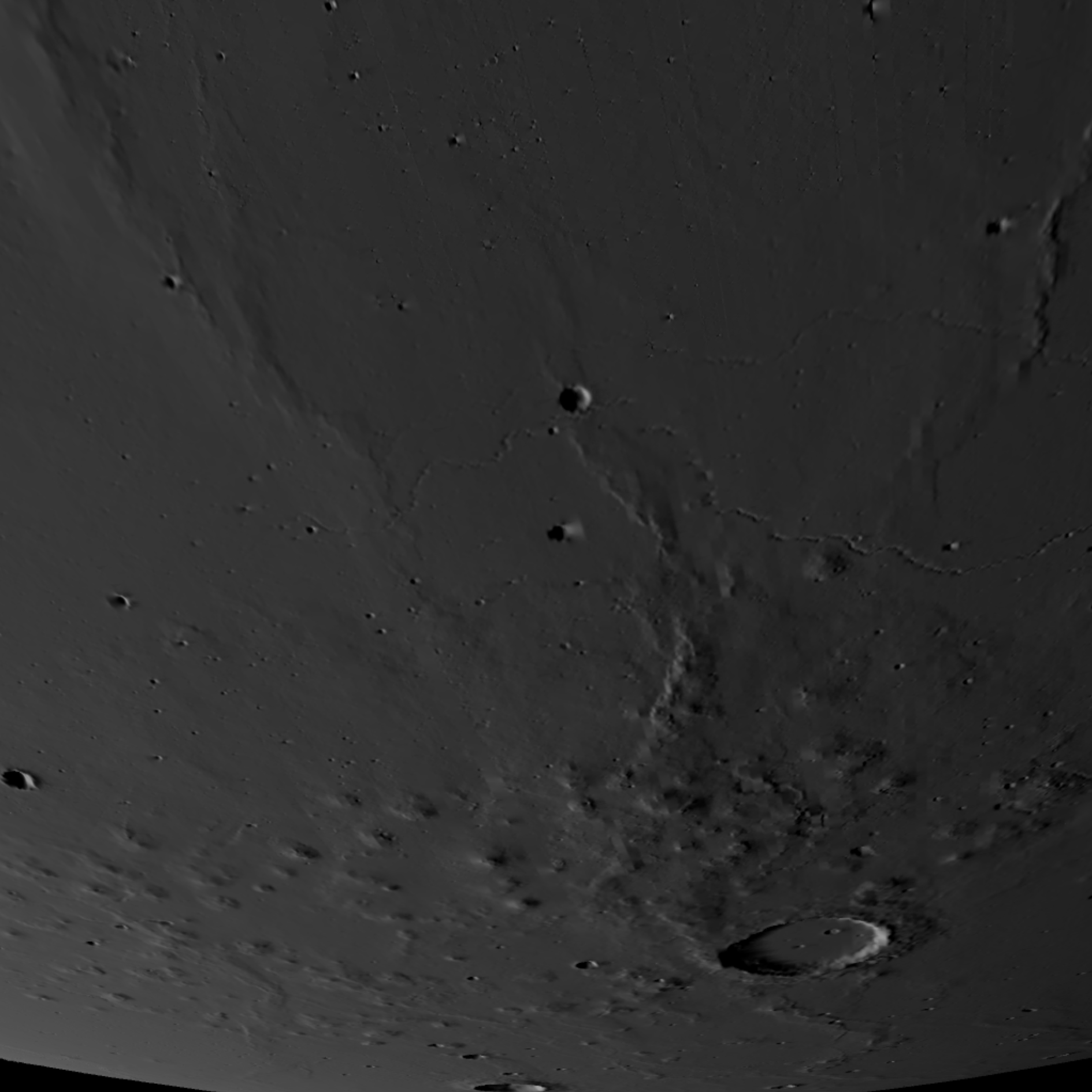}
        \caption{Oceanus Procellarum}
        \label{fig:167_oceanus_procellarum}
     \end{subfigure}
     \begin{subfigure}[b]{0.24\textwidth}
        \centering
        \includegraphics[width=\textwidth]{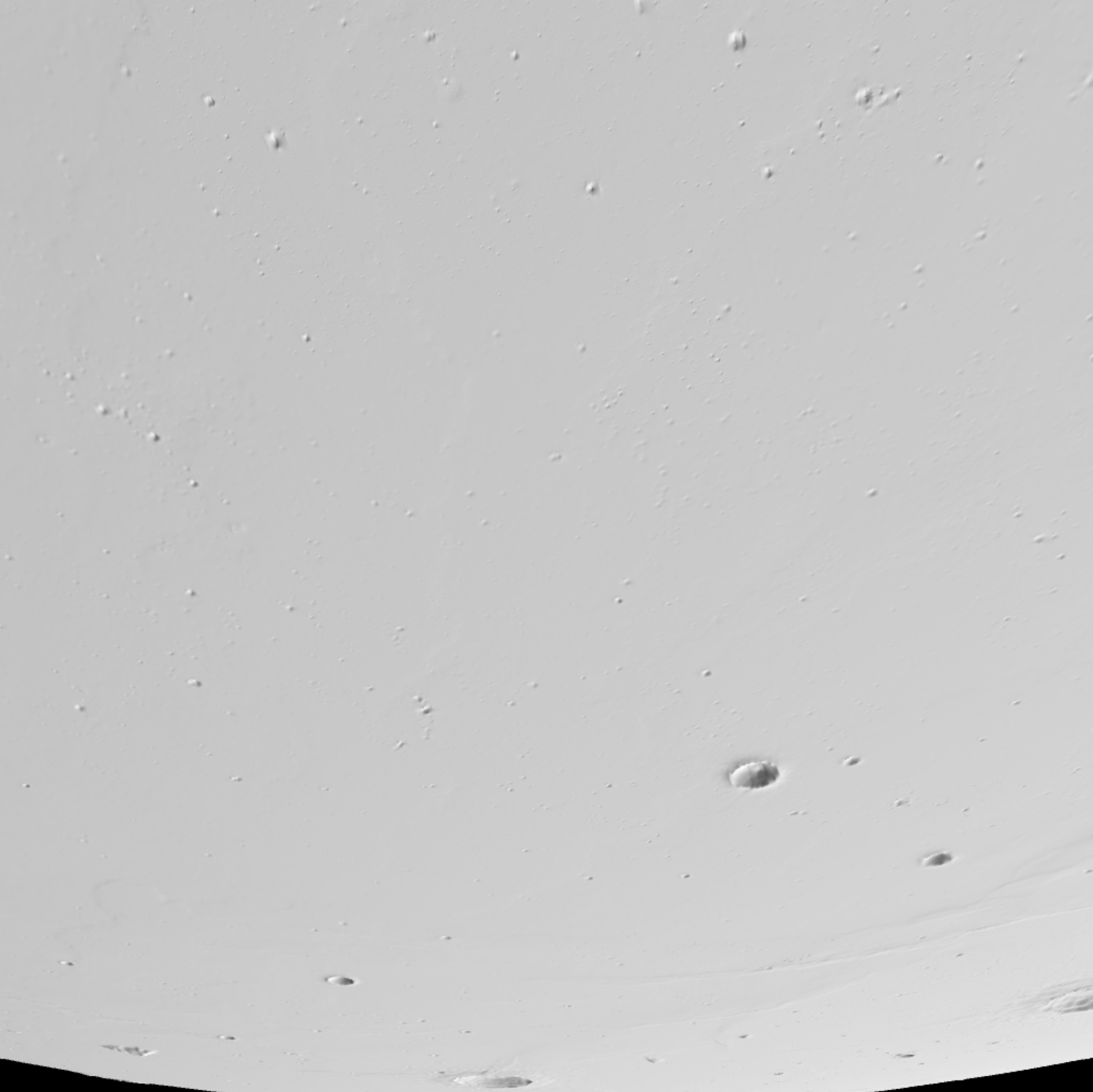}
        \caption{Mare Imbrium}
        \label{fig:4513_mare_imbrium}
     \end{subfigure}
     \caption{Four sample images taken from some of the Lunar Maria regions. These regions typically have few observable craters.}
     \label{fig:Lunar_Marina_regions}
\end{figure}

Another dataset attribute that will affect the performance of STELLA is the number of observable craters.  Fig.~\ref{fig:number_of_detected_craters} details the average number of craters detected by our CDA over CRESENT-365, visualised across a map of the lunar surface. From Fig.~\ref{fig:map_overlay}, we observe a strong correlation between lunar regions where there are fewer detected craters, and some of the Lunar Maria regions (see Fig.~\ref{fig:Lunar_Marina_regions}), which are known to be less cratered.

\subsection{Analysis of CDA}\label{sec:analysis_of_cda}

\begin{figure}
    \centering
    \includegraphics[width=0.6\linewidth]{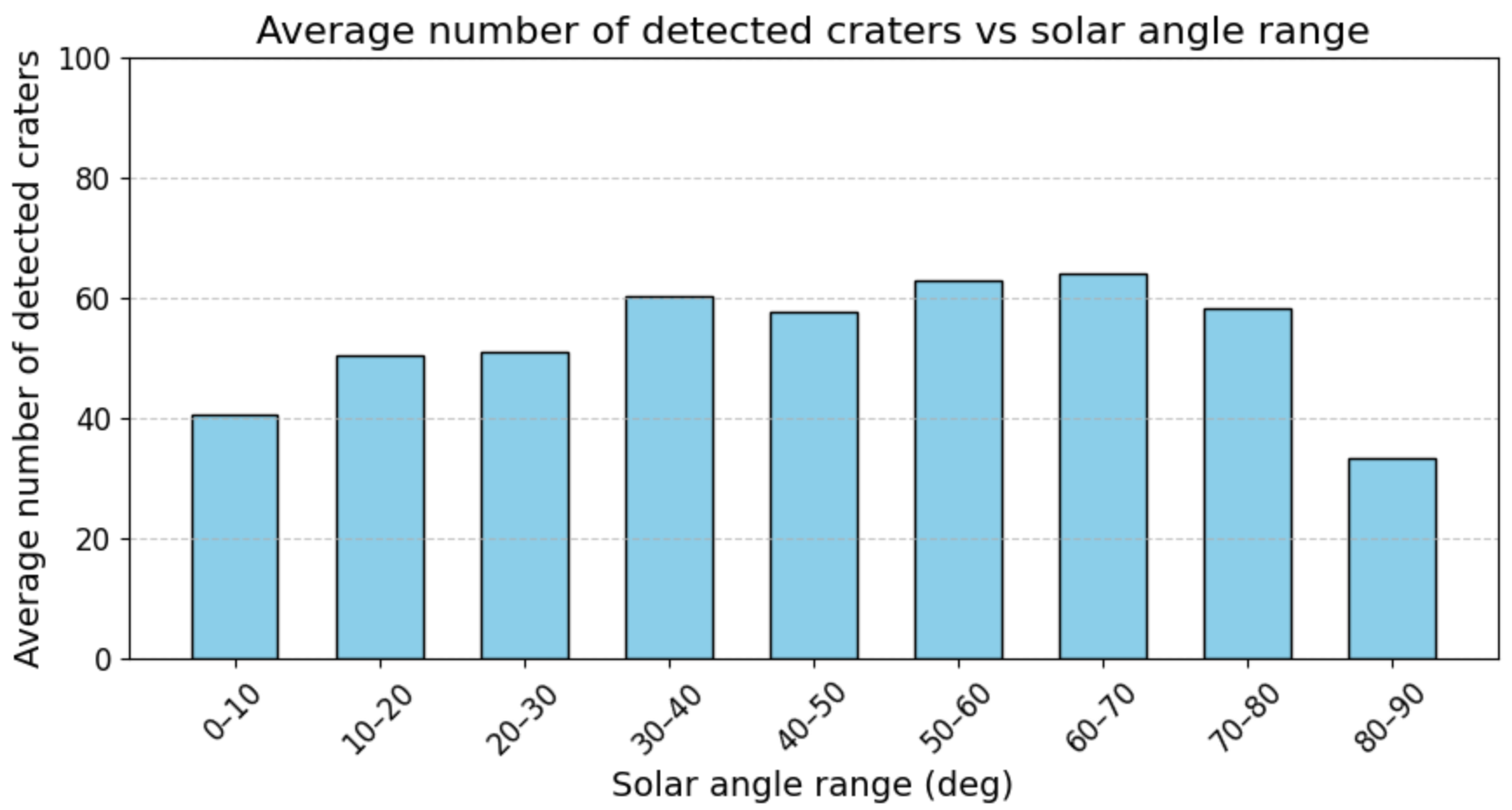}
    \caption{Graph of the average number of craters detected by the CDA for images taken within a solar angle range in the CRESENT-365 dataset.}
    \label{fig:num_craters_vs_solar_angle}
\end{figure}

\begin{figure}[htbp]
    \centering
     \begin{subfigure}[b]{0.3\textwidth}
         \centering
         \includegraphics[width=\textwidth]{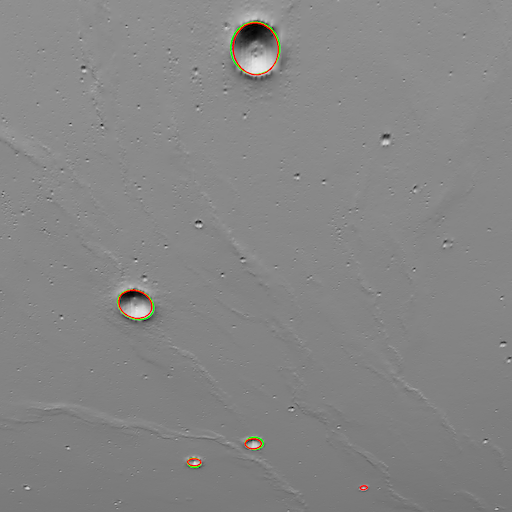}
         \caption{}
         \label{fig:cresent_prediction_1514}
     \end{subfigure}
     \begin{subfigure}[b]{0.3\textwidth}
         \centering
         \includegraphics[width=\textwidth]{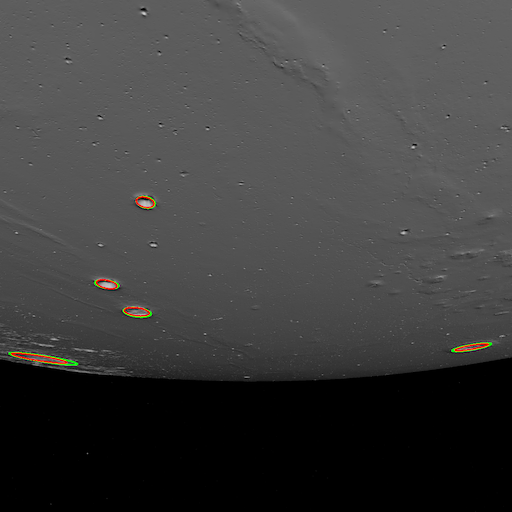}
         \caption{}
         \label{fig:cresent_prediction_833}
     \end{subfigure}
     \begin{subfigure}[b]{0.3\textwidth}
         \centering
         \includegraphics[width=\textwidth]{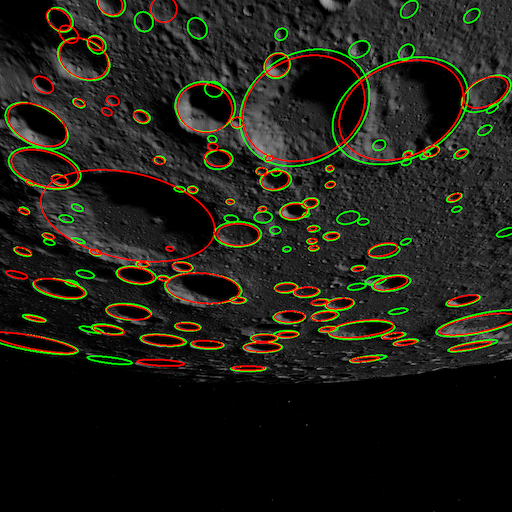}
         \caption{}
         \label{fig:cresent_prediction_804}
     \end{subfigure}
     \begin{subfigure}[b]{0.3\textwidth}
         \centering
         \includegraphics[width=\textwidth]{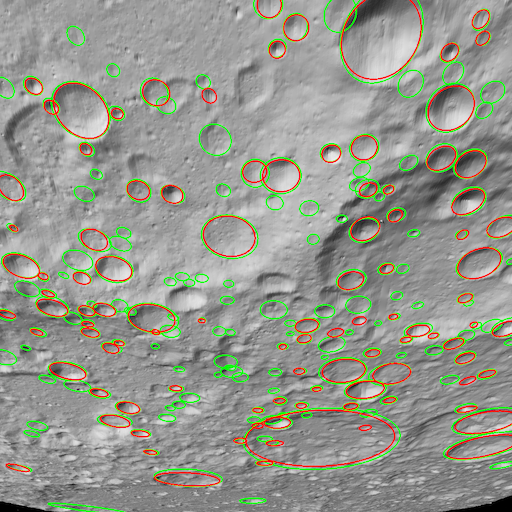}
         \caption{}
         \label{fig:cresent_prediction_535}
     \end{subfigure}
     \begin{subfigure}[b]{0.3\textwidth}
         \centering
         \includegraphics[width=\textwidth]{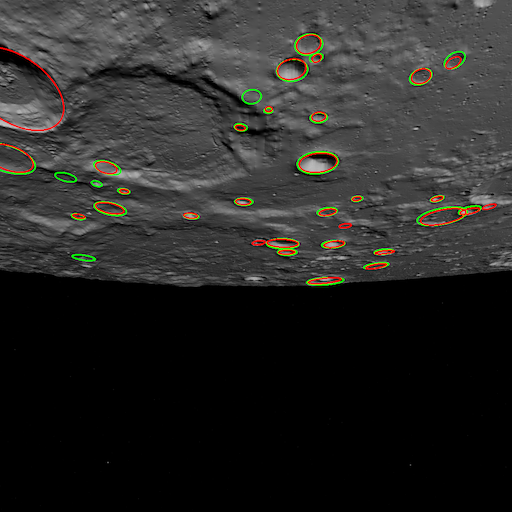}
         \caption{}
         \label{fig:cresent_prediction_61}
     \end{subfigure}
     \begin{subfigure}[b]{0.3\textwidth}
         \centering
         \includegraphics[width=\textwidth]{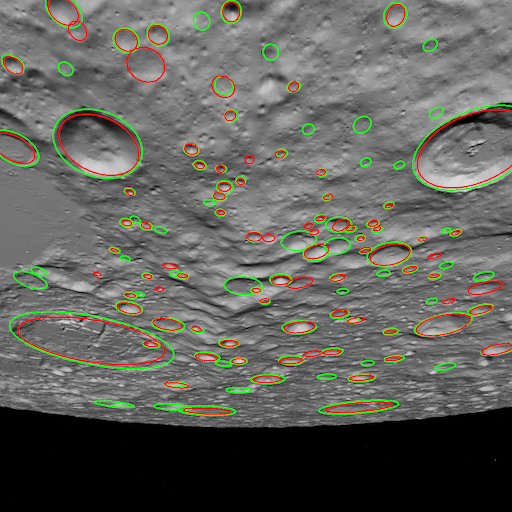}
         \caption{}
         \label{fig:cresent_prediction_1105}
     \end{subfigure}

     \begin{subfigure}[b]{0.3\textwidth}
         \centering
         \includegraphics[width=\textwidth]{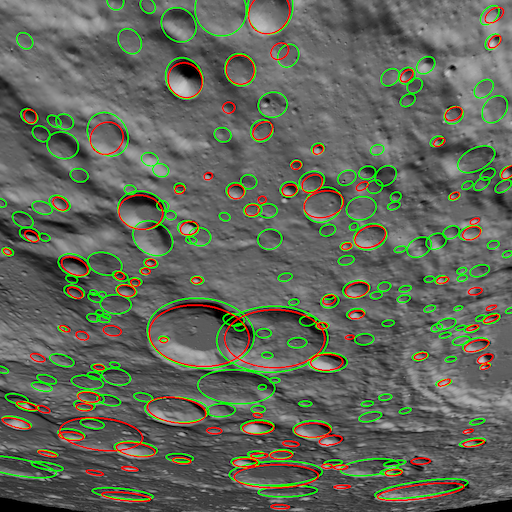}
         \caption{}
         \label{fig:orbital_predictions_10007}
     \end{subfigure}
     \begin{subfigure}[b]{0.3\textwidth}
         \centering
         \includegraphics[width=\textwidth]{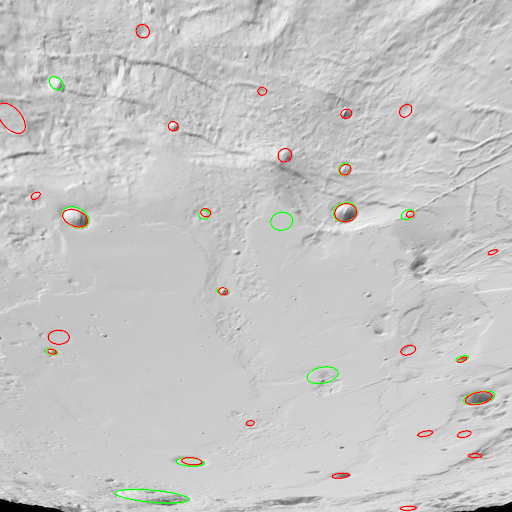}
         \caption{}
         \label{fig:orbital_predictions_10020}
     \end{subfigure}
     \begin{subfigure}[b]{0.3\textwidth}
         \centering
         \includegraphics[width=\textwidth]{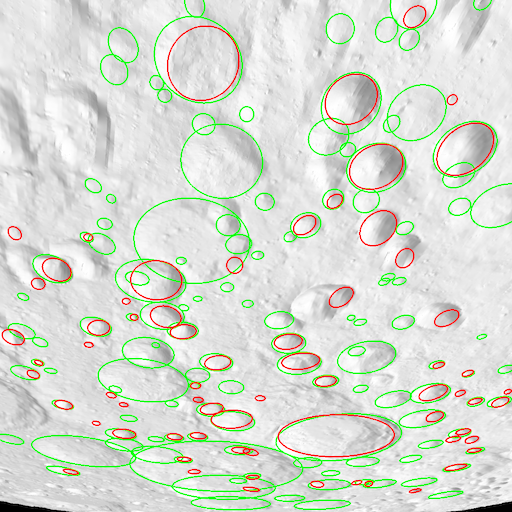}
         \caption{}
         \label{fig:orbital_predictions_10039}
     \end{subfigure}
     \begin{subfigure}[b]{0.3\textwidth}
         \centering
         \includegraphics[width=\textwidth]{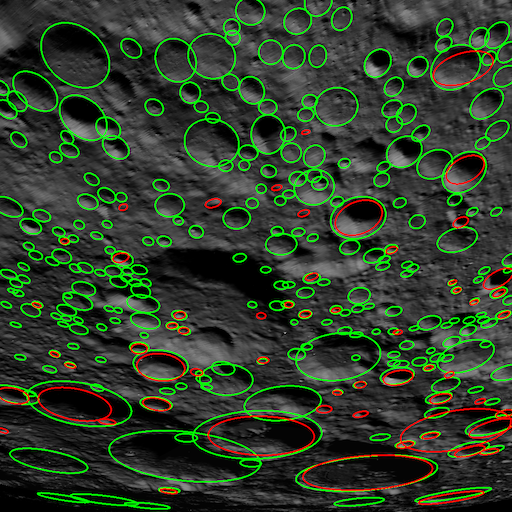}
         \caption{}
         \label{fig:orbital_predictions_10014}
     \end{subfigure}
     \begin{subfigure}[b]{0.3\textwidth}
         \centering
         \includegraphics[width=\textwidth]{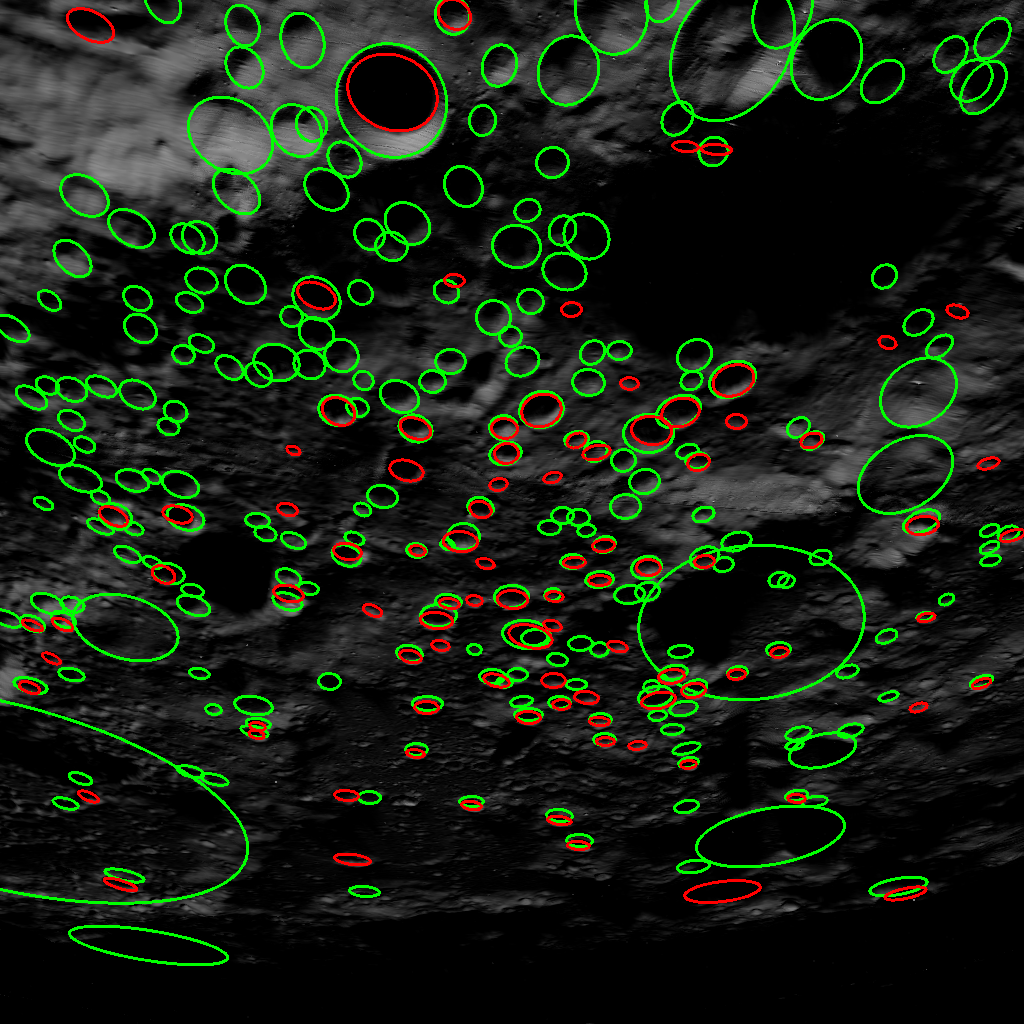}
         \caption{}
         \label{fig:orbital_predictions_2550}
     \end{subfigure}
     \begin{subfigure}[b]{0.3\textwidth}
         \centering
         \includegraphics[width=\textwidth]{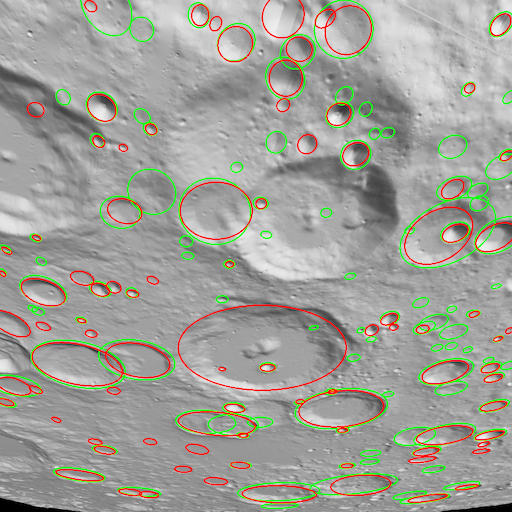}
         \caption{}
         \label{fig:orbital_predictions_10028}
     \end{subfigure}
    \caption{(a)-(f) CDA predictions on CRESENT+. (g)-(l) CDA predictions on CRESENT-365. Green ellipses are filtered ground truth craters, and red ellipses are predicted craters.}
    \label{fig:cda_predictions_orbital}
\end{figure}

Quantitative CDA results on the CRESENT+ and CRESENT-365 datasets can be seen in Tab.~\ref{tab:results} and Fig.~\ref{fig:num_craters_vs_solar_angle}, and qualitative results in Fig.~\ref{fig:cda_predictions_orbital}. 

\begin{table}[!hbt]
    \centering
    \renewcommand{\arraystretch}{1.2}
    \begin{tabular}{lcccccc}
        \toprule
        \textbf{Datset} & \textbf{Precision} & \textbf{Recall} & \textbf{F1 score} & \textbf{Avg. p} & \textbf{Avg. tp} \\
        \midrule
        CRESENT+ & 0.82 & 0.64 & 0.71 & 64.16 & 52.66 \\
        \midrule
        CRESENT-365 & 0.36 & 0.24 & 0.27 & 54.76 & 23.18 \\
        \bottomrule
    \end{tabular}
    
    \caption{CDA performance on CRESENT+ and CRESENT-365, considering precision, recall, F1 score, average number of detections (avg. p), average number of true positive detections (avg. tp).}
    \label{tab:results}
\end{table}

From these results, it is evident that the CDA is able to detect more craters on average with higher accuracy on the CRESENT+ test set than on CRESENT-365. This was expected, as CRESENT-365 introduces new lunar regions with various illumination conditions that could not be directly replicated through the augmentation of CRESENT+ during training. 

Fig.~\ref{fig:num_craters_vs_solar_angle} graphs the average number of craters detected by the CDA across a range of solar angles. This graph shows that the CDA can detect more craters on average for high contrast images, and detects the least number of craters when the input image is too dark,~\ie, in the solar angle ranges of $80$--$90^\circ$.

From the qualitative results in Fig.~\ref{fig:cda_predictions_orbital}, the CDA returns a subset of crater detections with no ground truth correspondence. We observe that some of these `false-positive detections' are real craters. These are the craters that were removed during the filtering process as detailed in Sec.~\ref{sec:cda}. The other \textit{legitimate} false positives will be handled by the outlier discarding component of CID and CBPE.

To improve CDA performance on the CRESENT-365 dataset, we suggest re-training or fine-tuning the CDA on a similar lunar mapping dataset to better account for the lighting conditions and terrain expected of CRESENT-365. However, we leave this as future work.

The following section evaluates the performance of STELLA, which is the central focus of this work.

\subsection{Performance of STELLA-core on CRESENT+}\label{sec:results_on_cresent}

In this section, we evaluate the STELLA-core pipeline on the test set of the CRESENT+. As CRESENT+ was not produced under an orbital trajectory, STELLA-OD was not evaluated. 


We obtained 1,773 pose estimates from STELLA-core, implying that no pose estimates could be returned for 146 images. This was expected as CRESENT+ contained images with no observable craters.


Tab.~\ref{tab:cresent_pose_results} contains the average, median, standard deviation (STD) and root mean square (RMS) errors of STELLA-core on the CRESENT+ test set. From these results, we observe much lower median observed surface and position errors than the corresponding average errors - a $70.5\%$ and $65.5\%$ reduction in median observed surface error and position error from the corresponding average error respectively.  This suggests that STELLA-core is robust to many instances of CRESENT+, with a subset of high-error outliers raising the mean, STD and RMS. The average, median and RMS angular error was $0.01^\circ$, implying that CBPE was also able to optimise attitude (recall that the error uncertainty of attitude was $0.02^\circ$).


\setlength{\extrarowheight}{2pt}

\begin{table}[hbt!]
    \centering
    \begin{minipage}[t]{0.9\linewidth}
        \centering
        \begin{tabularx}{\textwidth}{
            l
            >{\centering\arraybackslash}X
            >{\centering\arraybackslash}X
            >{\centering\arraybackslash}X
            >{\centering\arraybackslash}X
        }
            \toprule
            \textbf{STELLA-core} & \shortstack{observed\\surface error (m)} &
            \shortstack{position\\error (m)} & \shortstack{angular\\error (deg)} & \shortstack{runtime\\(s)} \\
            \midrule

            \textbf{Average} & 643.96 & 888.86 & 0.01 & 122.90 \\
            \textbf{Median}  & 189.97 & 315.89 & 0.01 & 71.28 \\
            \textbf{STD}  & 1830.41 & 1898.52 & 0.01 & 156.85 \\
            \textbf{RMS}  & 1939.90 & 2095.81 & 0.01 & 199.23 \\
            
            \bottomrule
        \end{tabularx}
        \caption{Performance of STELLA-core on CRESENT+.}
        \label{tab:cresent_pose_results}
    \end{minipage}
\end{table}

\begin{figure}[!hbt]
    \centering
    \includegraphics[width=0.8\linewidth]{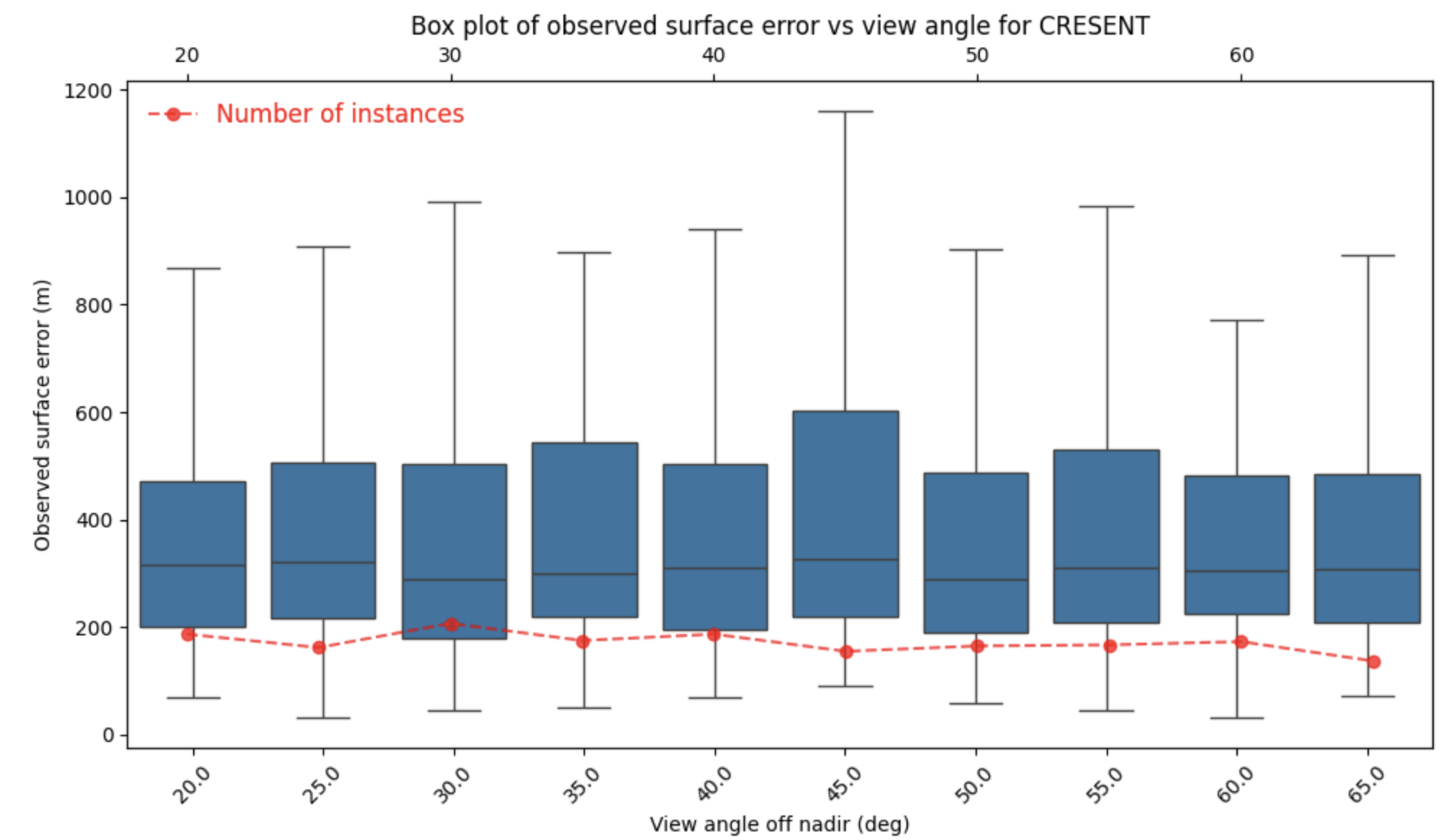}
    \caption{Performance of STELLA-core under different observation angles of CRESENT+ (fliers removed).}
    \label{fig:box_plot_angle_off_nad_cresent}
\end{figure}

As mentioned in Tab.~\ref{tab:landing_vs_surveillance}, lunar mapping missions are not limited to nadir viewing angles. CRESENT+ provides image data captured at viewing angles ranging from $20$--$65^\circ$ off nadir, enabling an evaluation of the robustness of STELLA-core under varying observation angles.  The box plot in Fig.~\ref{fig:box_plot_angle_off_nad_cresent} isolates the observed surface error of STELLA across each viewing angle.  Analysis of this box plot reveals a consistent median error across all viewing angles, indicating that STELLA-core is inherently robust to viewing perspective and does not observe noticeable deterioration in accuracy with increasing viewing angles. Considering this, STELLA could be extended to other operational scenarios where lunar surface imagery is acquired at non-nadir viewing angles - for instance, during trajectory adjustments in landing sequences.  In such cases, however, STELLA may need to be adapted to satisfy the associated mission performance constraints,~\eg, real-time pose estimation.


\subsection{Ablation study of STELLA-core on CRESENT-365}

This section presents the ablation study of the safeguarding mechanism incorporated into STELLA-core to reduce the likelihood of returning poor pose estimates. After generating a final pose, we evaluate the residuals~\eqref{eq:residual} for each catalogued–detected crater correspondence used by the CBPE component. To identify unreliable correspondences, we apply Tukey’s biweight loss function~\eqref{eq:tukey}, which was also used in CBPE, to flag correspondences with residuals exceeding the threshold $\alpha$ as outliers. The number of remaining inliers serves as a measure of pose quality, enabling STELLA-core to reject pose estimates with insufficient support.

\begin{figure}[!hbt]
    \centering
    \includegraphics[width=0.7\linewidth]{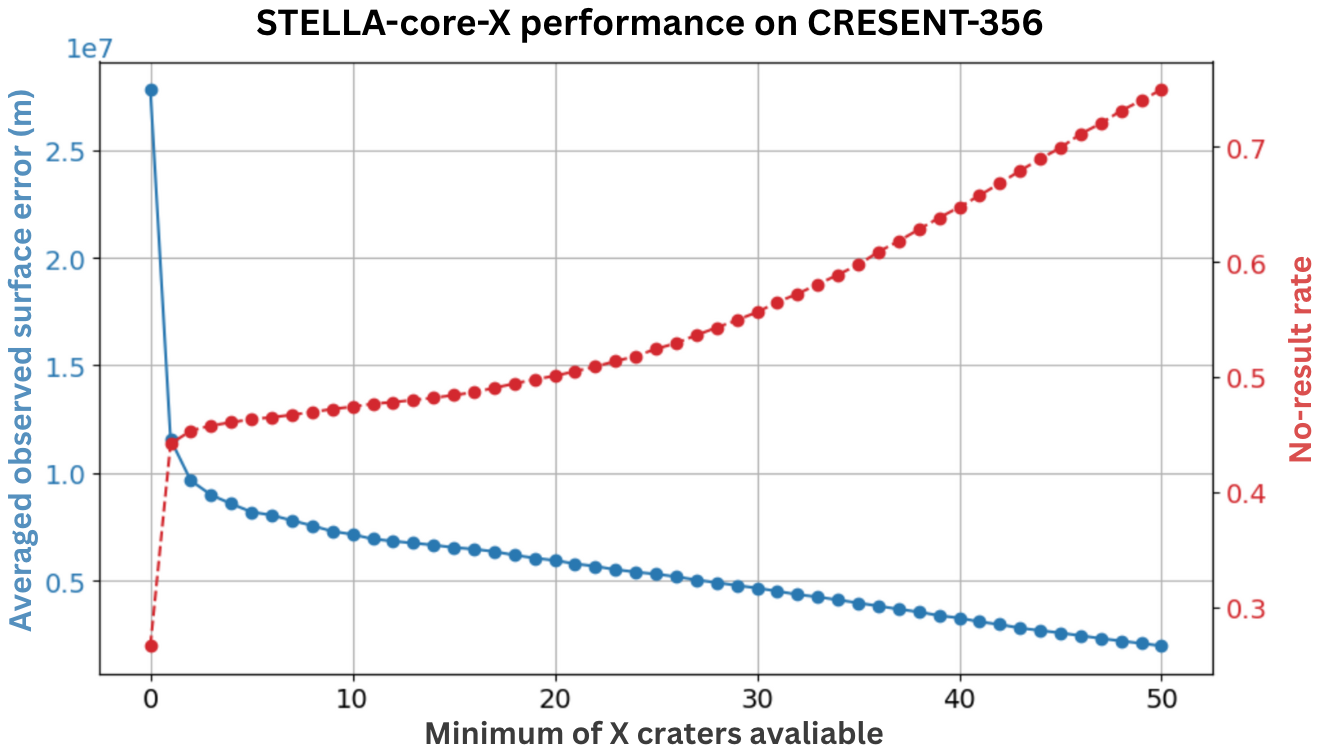}
    \caption{Average observed surface error and no-pose estimate result rate of STELLA-core-X on CRESENT-365.}
    \label{fig:STELLA_ablation}
\end{figure}

In Fig.~\ref{fig:STELLA_ablation}, the blue data points show the average observed surface error of STELLA-core-X on the CRESENT-365 dataset, where each average is computed over instances with at least X inlier correspondences. The red line indicates the number of instances in which STELLA-core-X fails to return a pose estimate. As shown, increasing the inlier threshold X leads to a lower average observed error, but at the cost of a higher no-result rate, reflecting a trade-off between accuracy and coverage.

Based on this ablation study, we additionally include \textbf{STELLA-core-6} in the evaluation on CRESENT-365. While using a higher inlier threshold can lead to improved average accuracy, we observe that the first performance plateau in observed surface error occurs at the six-correspondence mark. At this threshold, STELLA-core-6 still produces pose estimates for over 50\% of the data instances (see Fig.~\ref{fig:STELLA_ablation}), offering a balanced trade-off between accuracy and coverage.

\subsection{Performance of STELLA on CRESENT-365}\label{sec:cresent-365_results}
We discuss the performance of STELLA-core, STELLA-core-6 and STELLA-OD on the CRESENT-365 dataset. Based on the process described above, STELLA-core and STELLA-core-6 successfully returned a pose estimate for 11,213 and 8,179 data instances of the CRESENT-365 dataset respectively. We provide the estimates from STELLA-core-6 to STELLA-OD for the OD process. The best-fit orbit was then propagated to obtain a position estimate for the entire dataset, with 15,283 data instances.

The quantitative results of STELLA-core, STELLA-core-6 and STELLA-OD on the CRESENT-365 dataset can be seen in Tab.~\ref{tab:pose_results_orbital}.  Comparing these results to the ones presented in Tab.~\ref{tab:cresent_pose_results}, it is evident that CRESENT-365 provides a more challenging scenario than CRESENT+, producing higher error across all metrics (except angular error, which maintains an error of $0.01^\circ$) on average.

\begin{table}[hbt!]
    \centering
    \begin{minipage}[t]{0.9\linewidth}
        \centering
        \begin{tabularx}{\textwidth}{
            l  
            >{\centering\arraybackslash}X  
            >{\centering\arraybackslash}X  
            >{\centering\arraybackslash}X  
            >{\centering\arraybackslash}X  
        }
            \toprule
            \textbf{Method} & \shortstack{observed\\surface error (m)} &
            \shortstack{position\\error (m)} & \shortstack{angular\\error (deg)} & \shortstack{runtime\\(s)} \\
            \midrule

             \multicolumn{5}{l}{\textbf{Average}} \\
            \midrule
            STELLA-core & 2481.90 & 3012.93 & 0.01 & 154.81 \\
            STELLA-core-6 & 984.99 & 1572.27 & 0.01 & 207.48 \\
            STELLA-OD & 900.00 & 947.17 & - & - \\
            \addlinespace[2pt]

            \midrule
            \multicolumn{5}{l}{\textbf{Median}} \\
            \midrule
            STELLA-core & 534.22 & 750.79 & 0.01 & 89.69 \\
            STELLA-core-6 & 355.76 & 512.50 & 0.01 & 124.91 \\
            STELLA-OD & 757.21 & 755.65 & - & - \\
            \toprule

            \multicolumn{5}{l}{\textbf{STD}} \\
            \midrule
            STELLA-core & 3447.08 & 3658.46 & 0.01 & 233.15 \\
            STELLA-core-6 & 1980.53 & 2870.24 & 0.01 & 251.86 \\
            STELLA-OD & 551.52 & 414.79 & - & - \\
            \addlinespace[2pt]
            \midrule
            
            \multicolumn{5}{l}{\textbf{RMS}} \\
            \midrule
            STELLA-core & 4247.45 & 4739.19 & 0.01 & 279.80 \\
            STELLA-core-6 & 2210.84 & 3271.66 & 0.01 & 326.22 \\
            STELLA-OD & 1055.55 & 1034.01 & - & - \\
            \addlinespace[2pt]
            \bottomrule
        \end{tabularx}
        \caption{Results on CRESENT-365.}
        \label{tab:pose_results_orbital}
    \end{minipage}
\end{table}


From Tab.~\ref{tab:pose_results_orbital}, STELLA-OD has the lowest error average across all evaluation metrics, observing a 63.7\% reduction in the average observed surface error from STELLA-core and an 8.6\% reduction from STELLA-core-6. STELLA-OD also observes a 68.6\% reduction in average position error from STELLA-core and a 39.6\% reduction from STELLA-core-6.  Furthermore, STELLA-OD achieves the lowest RMS error across all evaluation metrics, observing a 75.15\% reduction in the RMS observed surface error from STELLA-core and a 52.26\% reduction from STELLA-core-6. STELLA-OD also observes a 78.18\% reduction in RMS position error from STELLA-core and a 68.39\% reduction from STELLA-core-6.

STELLA-core-6 achieves the lowest median error across all evaluation metrics, observing a 53.0\% reduction in the median observed surface error from STELLA-OD and a 33.4\% reduction from STELLA-core. STELLA-core-6 also observes a 32.2\% reduction in median position error from STELLA and a 31.7\% reduction from STELLA-core.

\begin{figure}[!hbt]
    \centering
    \includegraphics[width=0.80\linewidth]{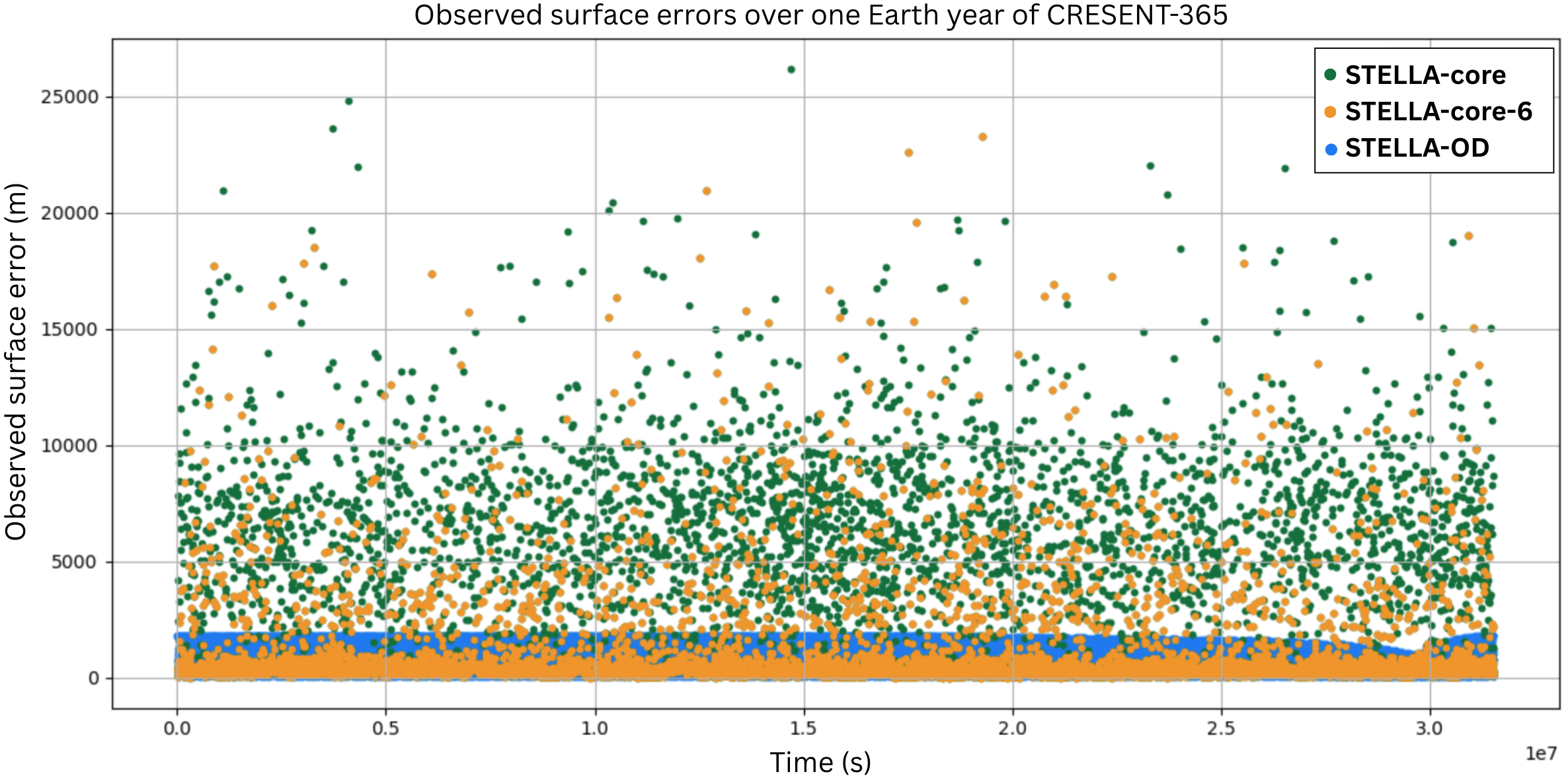}
    \caption{Performance of STELLA-core, STELLA-core-6 and STELLA over one Earth year.}
    \label{fig:stella_time_performance}
\end{figure}

From these results, it is evident that STELLA-OD is more stable, yielding the lowest STD observed surface and position errors compared to STELLA-core and STELLA-core-6.  The stability of STELLA-OD stems from its least-squares orbit-fitting component: each position is derived from a single trajectory that is optimally fitted to all positions provided by STELLA-core-6. As the fit averages over many positions, an occasional poor estimate cannot dominate the solution. This behaviour is also exhibited in Fig.~\ref{fig:stella_time_performance}, where STELLA-OD produces stable observed surface error results in the range of 46.22--1,793.36~m. Meanwhile, STELLA-core and STELLA-core-6 have a high concentration of lower observed surface error results that reflect the median error in Tab.~\ref{tab:pose_results_orbital}. However, these variants are also susceptible to producing high-error estimates, sometimes exceeding 25,000m+. 



The results in Fig.~\ref{fig:stella_time_performance} suggest a potential enhancement to the STELLA pipeline. Given that STELLA-OD exhibits a relatively predictable error profile, pose estimates from STELLA-core could be post-validated by checking whether they fall within the expected error bounds of the corresponding STELLA-OD prediction. If this condition is met, the STELLA-core estimate is retained (since it has a better median error performance than STELLA-OD); otherwise, it is discarded. Such a filtering strategy could effectively eliminate large-error estimates from STELLA-core, thereby improving the overall average accuracy of the pipeline. While this approach shows promise, we leave its integration and evaluation as future work.

The superiority of our CBN approach over non-vision-based navigation (e.g., radio ranging) is also evident in Tab.~\ref{tab:pose_results_orbital}. As discussed in Sec.~\ref{sec:cid}, radio ranging yields an average position error of approximately $6.7 \mathrm{km}$ - exceeding the $3\sigma$ bound of STELLA-OD ($\mu = 0.94km, 3\sigma = 3(0.41km)$). Moreover, because CBN does not rely on continuous communications, it remains effective under degraded or denied links, making it a promising direction for navigation systems.

\begin{figure}[!hbt]
    \centering
     \begin{subfigure}[b]{0.38\textwidth}
        \centering
        \includegraphics[width=\textwidth]{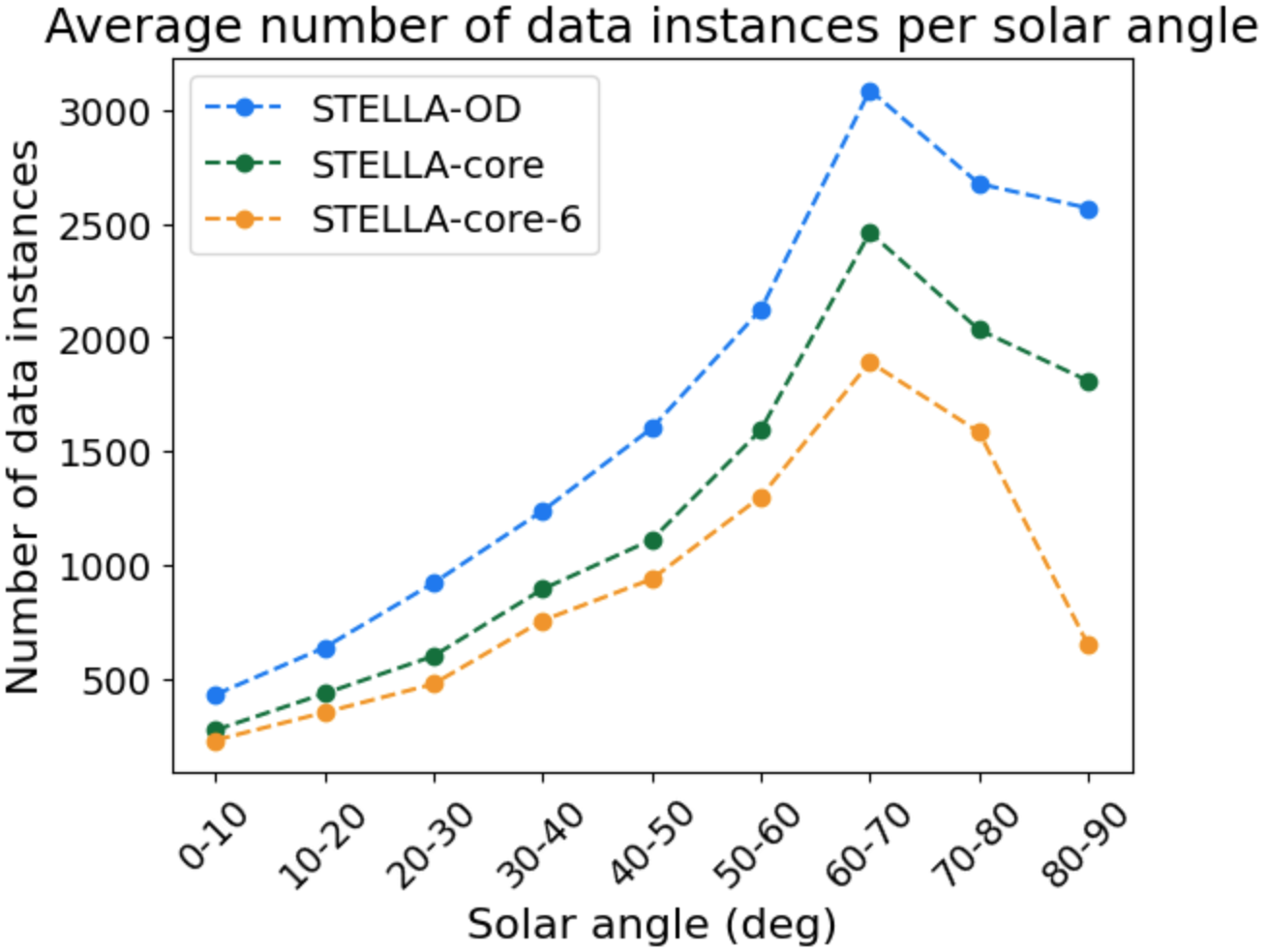}
        \caption{}
        \label{fig:solar_angle_before_after_filtering}
     \end{subfigure}
     \begin{subfigure}[b]{0.6\textwidth}
        \centering
        \includegraphics[width=\textwidth]{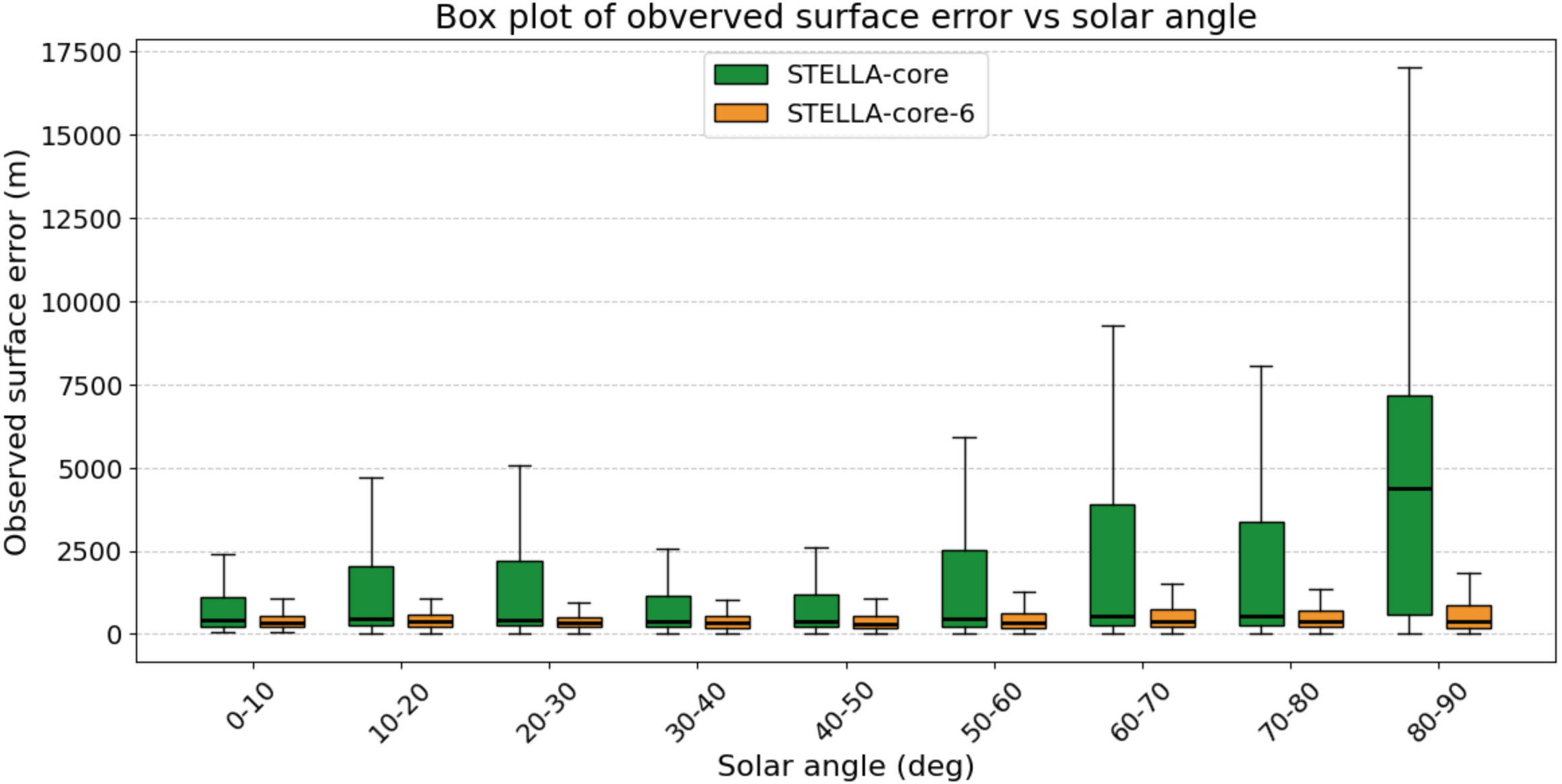}
        \caption{}
        \label{fig:box_plot_of_surface_error_vs_solar_angle}
     \end{subfigure}
     \label{}
     \caption{(a) Number of data instances per solar angle that returned a pose estimate result by STELLA-OD, STELLA-core and STELLA-core-6. (b) Box plots of observed surface error per solar angle for STELLA-core and STELLA-core-6 (note that fliers have been removed).}
\end{figure}

Fig.~\ref{fig:solar_angle_before_after_filtering} plots the number of processed data instances at each solar angle for STELLA-core, STELLA-core-6 and STELLA-OD.  It is evident that as the solar angle increases, the number of data instances processed by STELLA-core and STELLA-core-6 decreases. STELLA-core-6 observes the largest decrease of 1,161 processed data instances from STELLA-core, in the solar angle range of $80$--$90^\circ$.  This aligns with the findings from Fig.~\ref{fig:num_craters_vs_solar_angle}, where the solar angle range of $80$--$90^\circ$ yielded the lowest average number of crater detections by the CDA.

From the data instances processed by STELLA-core and STELLA-core-6, a box plot on the observed surface error per solar illumination angle range can be seen in Fig.~\ref{fig:box_plot_of_surface_error_vs_solar_angle}. While the performance of STELLA-core shows higher variability compared to STELLA-core-6 across the range of solar angles, the median error is reasonably constant from $0$--$80^\circ$. Under the solar angles of $80$--$90^\circ$, STELLA-core does observe a large increase in observed surface error. Comparatively, the median error produced by STELLA-core-6 appears constant across all solar angles, indicating that the data instances contributing to the large STELLA-core error in the solar angles of $80$--$90^\circ$ were discounted in STELLA-core-6 (which is reflective of the findings in Fig.~\ref{fig:solar_angle_before_after_filtering}).  From these results, we can conclude that the STELLA-core pipeline is generally robust to solar angles within the visible range, with a potential increase in error under very low illumination conditions ($80$--$90^\circ$ solar angle).

\begin{figure}[!hbt]
    \centering
     \begin{subfigure}[b]{0.48\textwidth}
         \centering
        \includegraphics[width=\textwidth]{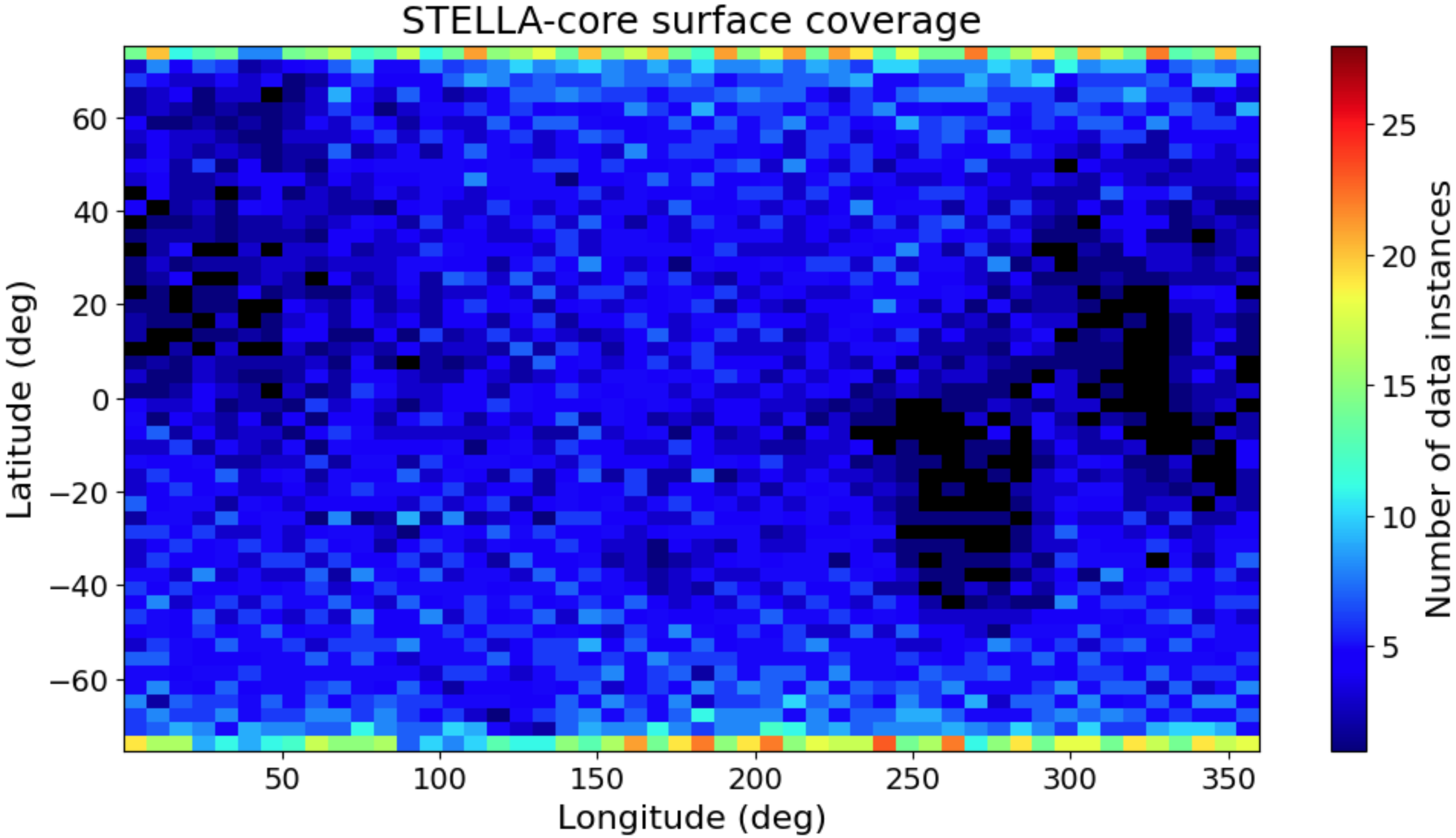}
        \caption{}
        \label{fig:stella_coverage}
     \end{subfigure}
     \begin{subfigure}[b]{0.48\textwidth}
         \centering
        \includegraphics[width=\textwidth]{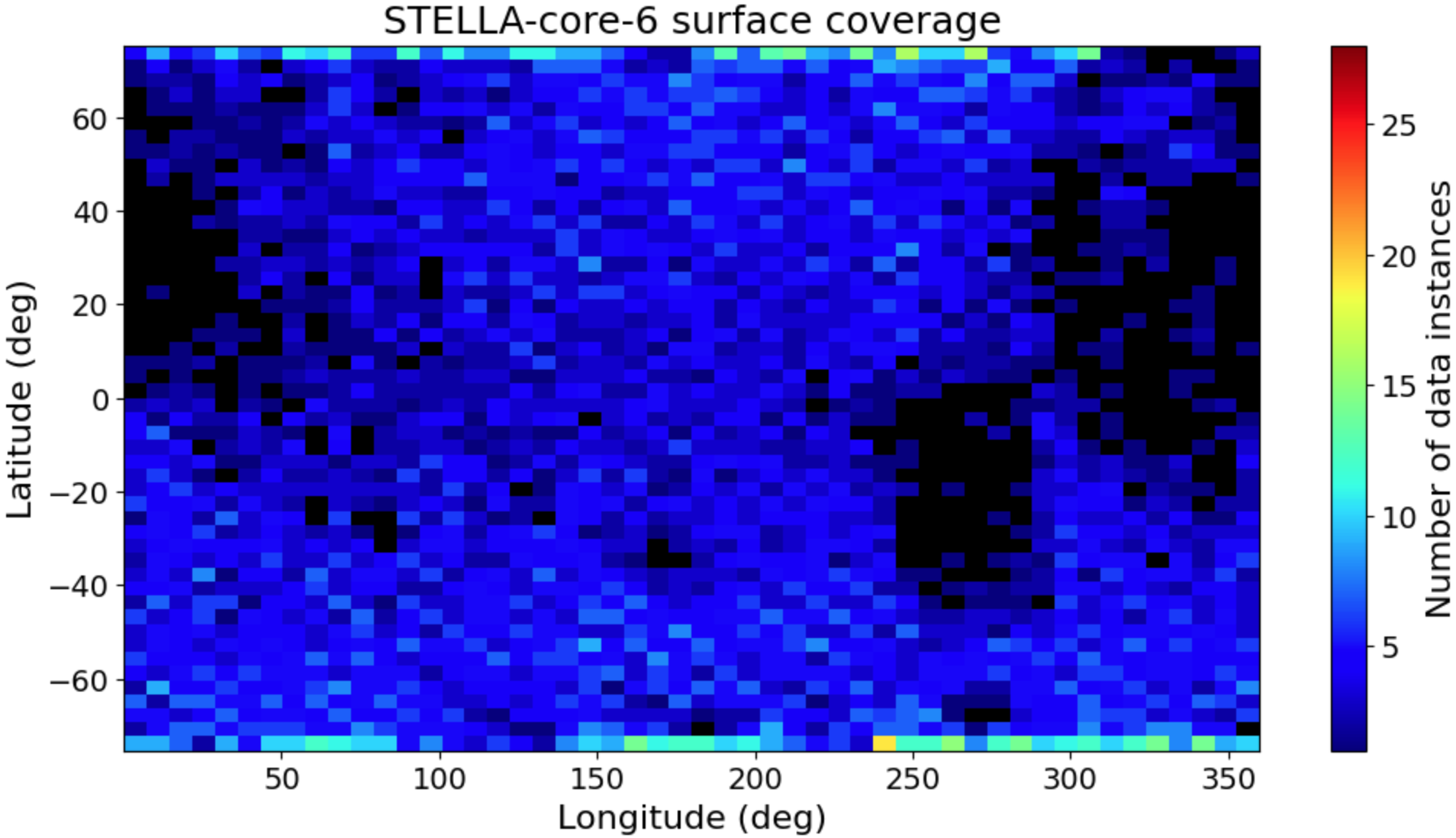}
        \caption{}
        \label{fig:stella_x_coverage}
     \end{subfigure}
     \caption{Number of data instances that returned a pose result for (a) STELLA-core and (b) STELLA-core-6, visualised across the lunar surface.}
     \label{fig:stellas_coverage}
\end{figure}

While the CRESENT-365 dataset has near-global coverage (see Fig.~\ref{fig:map_coverage}), the surface coverage after STELLA-core and STELLA-core-6 results in regions of the lunar surface where no predictions were made can be seen in Fig.~\ref{fig:stellas_coverage}. These empty regions have a strong correlation to the corresponding lunar regions that have fewer CDA-detected craters from Fig.~\ref{fig:number_of_detected_craters}.  From these results, we have identified the lunar regions that are challenging for optical navigation systems to make accurate pose estimates, informing future missions on where is best to take images and where to avoid.

\begin{figure}[!hbt]
    \centering
     \begin{subfigure}[b]{0.48\textwidth}
        \centering
        \includegraphics[width=\textwidth]{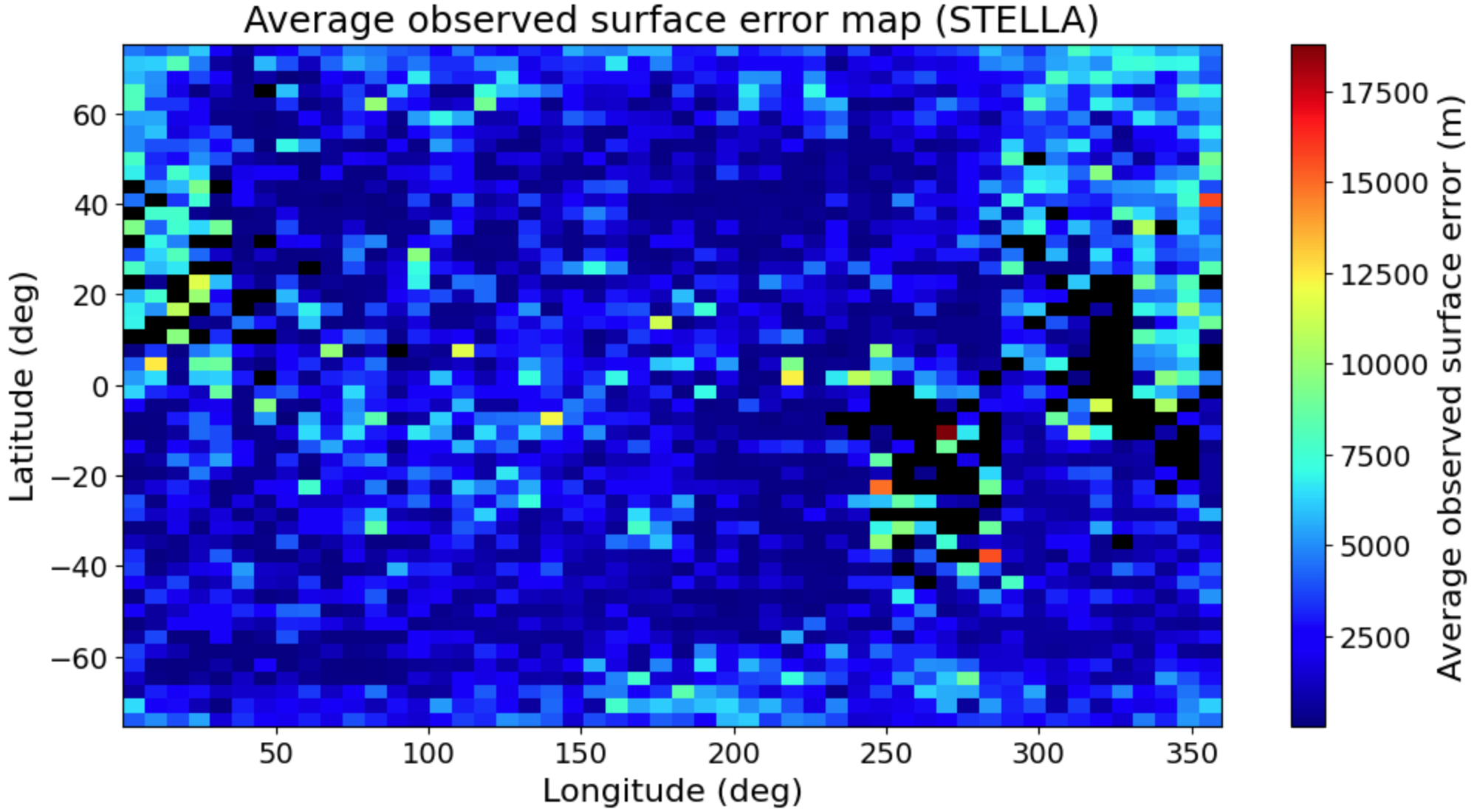}
        \caption{}
        \label{fig:stella_error}
     \end{subfigure}
     \begin{subfigure}[b]{0.48\textwidth}
        \centering
        \includegraphics[width=\textwidth]{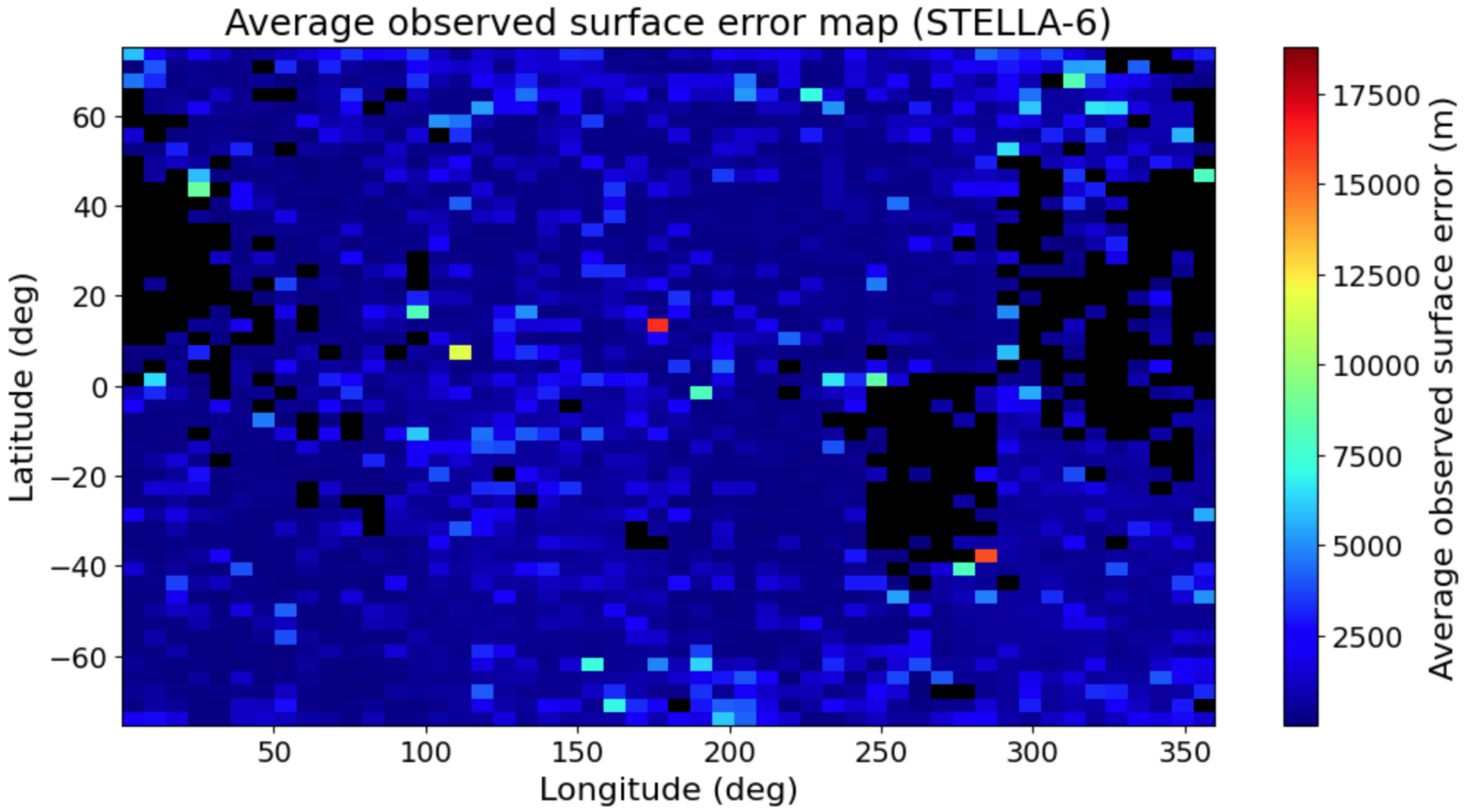}
        \caption{}
        \label{fig:stella-x_error}
     \end{subfigure}
     \caption{Average observed surface error visualised on a map of the lunar surface for (a)  STELLA-core, and (b) STELLA-core-6.}
     \label{fig:map_of_stella_errors}
\end{figure}

Fig.~\ref{fig:map_of_stella_errors} maps the average observed surface error of STELLA-core and STELLA-core-6 visualised across the corresponding observed lunar surface regions.  From these figures, we observe STELLA-core having larger errors around the regions where no pose estimates were made. Comparatively, as the boundaries of these regions have no pose estimate made by STELLA-core-6, the map of STELLA-core-6 demonstrates a more uniform error over the entire lunar surface, reinforcing the earlier finding that data instances containing a higher number of inlier crater correspondences yield more accurate pose estimates on average (Fig.~\ref{fig:STELLA_ablation}).

In this section, we have evaluated the performance of STELLA on the CRESENT+ test set and CRESENT-365 dataset. From these results, we have demonstrated that our proposed STELLA pipeline is robust to camera viewing angle and most surface illumination conditions.  Through an ablation study of STELLA-core, we observe a correlation between having a larger number of inlier crater correspondences with lower observed surface error, which provides a gauge for pose estimate quality.  With the integration of OD, the STELLA pipeline was able to provide pose estimates with metre-level position accuracy and sub-degree attitude accuracy on average across the 15,283 images from the year-long CRESENT-365 lunar mapping mission. Furthermore, the STELLA-core pipeline has identified regions of the lunar surface (namely, the Lunar Maria regions) where CBN is unreliable, informing future missions to avoid taking images in these regions if operationally viable.

\section{Conclusion}\label{sec:conclusion}
In this paper, we introduced STELLA, a novel Crater-Based Navigation (CBN) pipeline tailored specifically for long-duration lunar mapping missions. We also presented CRESENT-365, a comprehensive and realistic synthetic dataset emulating a year-long lunar orbital campaign, spanning a broad range of illumination conditions, global lunar surface coverage, and oblique viewing angles. Through extensive evaluations conducted on both the CRESENT+ and CRESENT-365 datasets, STELLA consistently demonstrated robustness across varying viewing angles and most lighting scenarios, achieving, on average, metre-level position accuracy and sub-degree attitude accuracy. As the first work that thoroughly assesses the applicability of CBN to lunar mapping scenarios, the performance of STELLA serves as a promising baseline, and lays the groundwork for further developments and deployments of vision-based navigation solutions for long-duration lunar orbital missions. Additionally, our evaluations identified specific lunar regions with challenging terrain and sparse crater features, providing valuable insights for planning future lunar mapping missions.

\section*{Acknowledgements}
This work was supported by the Australian Research Council and SmartSat CRC.

\section*{Declaration of competing interest}
The authors have no competing interests to declare that are relevant to the content of this article.

\section*{References}



\bibliographystyle{astrobib}
\vspace*{-1em}
\subsection*{Author biography}


\begin{biography}[author1]{Sofia McLeod} received her B.CompSci(Adv) (2019) and B.CompSci(Hons) (2020) degrees from the University of Adelaide, Australia. She is currently pursuing a Ph.D. degree at the Australian Institute for Machine Learning (AIML), University of Adelaide, Australia. Her research interests include vision-based navigation systems for spacecraft. E-mail: sofia.mcleod@adelaide.edu.au
\end{biography}

\begin{biography}[author2]{Chee-Kheng Chng} is a Senior Research Associate at the Australian Institute for Machine Learning (AIML) at the University of Adelaide. He obtained his Ph.D. in Computer and Mathematical Sciences from the University of Adelaide in 2023. His research focuses on solving problems in the space domain by leveraging advanced AI and geometry optimisation techniques. He has published on these topics in journals such as IEEE Transactions on Aerospace and Electronic Systems (TAES), Acta Astronautica, and Monthly Notices of the Royal Astronomical Society (MNRAS). E-mail: cheekheng.chng@adelaide.edu.au
\end{biography}

\vspace*{-1em}
\subsection*{Graphical table of contents}

\begin{figure}[!hbt]
    \centering
    \includegraphics[width=0.8\linewidth]{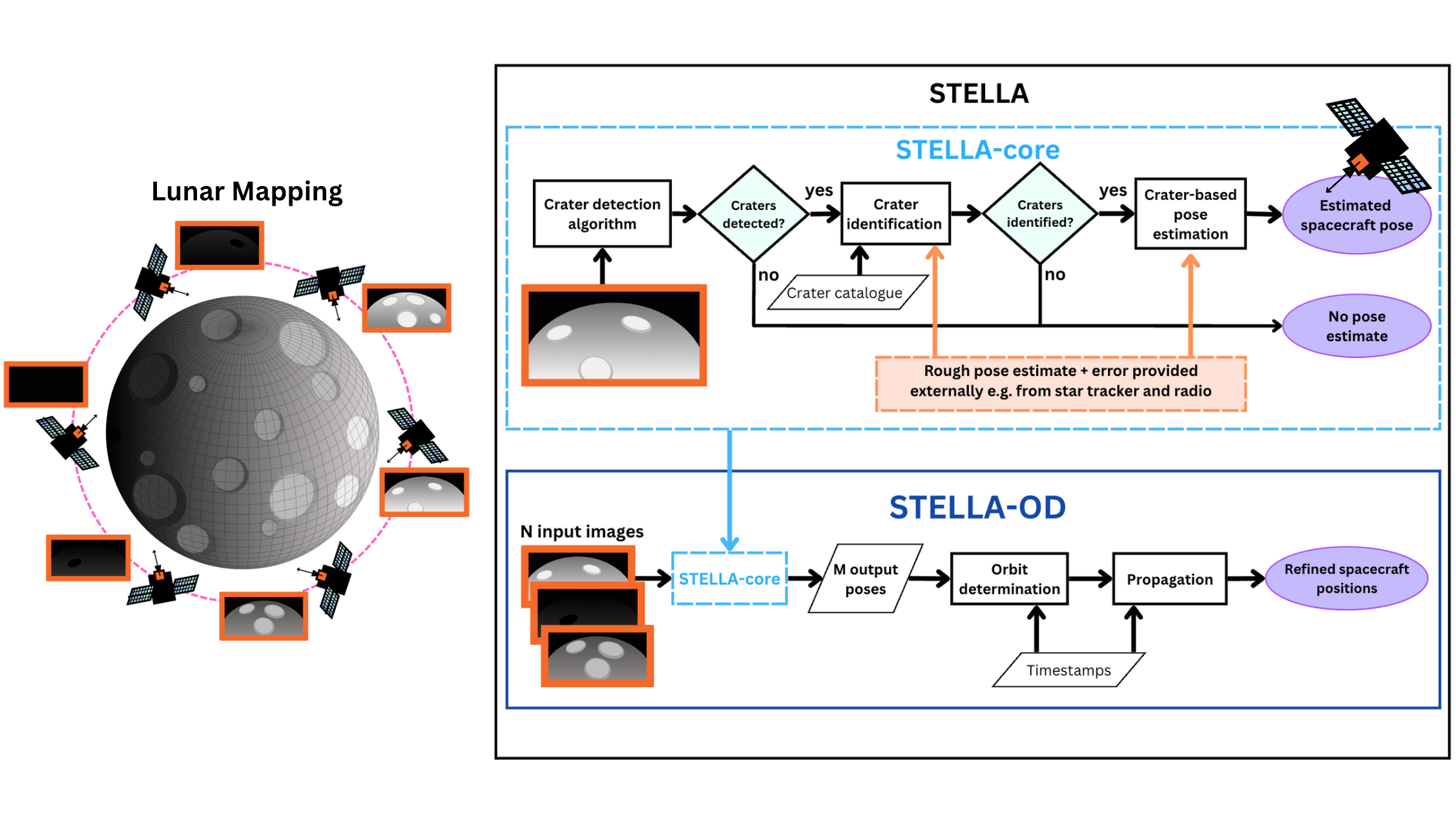}
    \caption{This work proposes STELLA, an AI-enabled crater-based navigation pipeline for lunar mapping missions.}
    \label{fig:graphical_table_of_contents}
\end{figure}


\section*{Supplementary material}
\section{Deriving ellipse parameters from a conic matrix}\label{app:ellipse_derivation}

The ellipse parameters $\mathcal{E} = \{x, y, \hat{a}, \hat{b}, \theta\}$ can be derived from conic $\mathbf{Q}$,

\begin{equation}
    \mathbf{Q} = 
    \begin{bmatrix}
    A & B & D \\
    B & C & E \\
    D & E & F 
    \end{bmatrix},
\end{equation}

where,

\begin{equation}
    x = (2CD-BE)/(B^2-4AC),
\end{equation}
\begin{equation}
    y = (2AE-BD)/(B^2-4AC),
\end{equation}
\begin{equation}
    \hat{a} = \dfrac{\sqrt{(2(AE^2+CD^2 - BDE + F(B^2-4AC))}}{(B^2-4AC)(\sqrt{(A-C)^2+B^2}-A-C))},
\end{equation}
\begin{equation}
     \hat{b} = \dfrac{\sqrt{(2(AE^2+CD^2 - BDE + F(B^2-4AC))}}{(B^2-4AC)(-\sqrt{(A-C)^2+B^2}-A-C))},
\end{equation}
\begin{equation}\label{eq:ellipse_error}
    \theta = 
    \begin{cases}
        \pi/2 &\text{ if } B = 0 \text{ and } A > C \\
        \dfrac{\cot^{-1}((A-C)/B)}{2} &\text{ if } B \neq 0 \text{ and } A \leq C \\
        \pi/2+\dfrac{\cot^{-1}((A-C)/B)}{2} &\text{ if } B \neq 0 \text{ and } A > C \\
        0 & \text{otherwise}
    \end{cases}.
\end{equation}

\end{document}